\documentclass[twoside]{article}

\usepackage[accepted]{aistats2026} %

\usepackage[utf8]{inputenc} %
\usepackage[T1]{fontenc}    %
\usepackage{hyperref}       %
\usepackage{url}            %
\usepackage{booktabs}       %
\usepackage{amsfonts}       %
\usepackage{amsthm}
\usepackage{nicefrac}       %
\usepackage{microtype}      %
\usepackage{xcolor}         %
\usepackage{comment}
\usepackage{amsmath}
\usepackage[inline]{enumitem} %
\usepackage{algorithm}
\usepackage{algpseudocode}
\usepackage{amsmath}
\usepackage{amssymb}
\usepackage{graphicx}
\usepackage{subcaption}     %
\usepackage{cleveref}       %
\usepackage{placeins}       %
\usepackage{tikz}           %
\usepackage{multicol}       %
\usepackage{microtype}      %
\usepackage{makecell}
\usepackage{float}
\usepackage{adjustbox}
\usepackage{accessibility}

\definecolor{gold}{RGB}{221, 196, 65}
\definecolor{silver}{RGB}{215, 215, 215}
\definecolor{bronze}{RGB}{126, 66, 5}

\newcommand{\tikzcircle}[2][red,fill=red]{\tikz[baseline=-0.7ex]\draw[#1,radius=#2] (0,0) circle ;}%

\newcommand{\textBF}[1]{%
  \pdfliteral direct {2 Tr 0.7 w} %
  #1%
  \pdfliteral direct {0 Tr 0 w}%
}

\newcommand{\gm}{
  \tikzcircle[gold,fill=gold]{2pt}
}
\newcommand{\sm}{
  \tikzcircle[silver,fill=silver]{2pt}
}
\newcommand{\bm}{
  \tikzcircle[bronze,fill=bronze]{2pt}
}

\newcommand{\loss}{\mathcal{L}}

\newcommand{\hy}{\hat{y}}

\usepackage[
  backend=biber,
  style=authoryear,
  maxcitenames=1,
  mincitenames=1,
  maxbibnames=99,
  uniquelist=false,
  uniquename=false
]{biblatex}
\begin{document}

\twocolumn[
  \aistatstitle{
    RamPINN: Recovering Raman Spectra From Coherent Anti-Stokes Spectra Using Embedded Physics
  }
  \aistatsauthor{
    Sai Karthikeya Vemuri \quad
    Adithya Ashok Chalain Valapil \quad
    Tim Büchner \quad
    Joachim Denzler
  }
  \aistatsaddress{
    Computer Vision Group, Friedrich Schiller University Jena, Germany
  }
]

\begin{abstract}
  Transferring the recent advancements in deep learning into scientific disciplines is hindered by the lack of the required large-scale datasets for training.
  We argue that in these knowledge-rich domains, the established body of scientific theory provides reliable inductive biases in the form of governing physical laws.
  We address the ill-posed inverse problem of recovering Raman spectra from noisy Coherent Anti-Stokes Raman Scattering (CARS) measurements, as the true Raman signal here is suppressed by a dominating non-resonant background.
  We propose RamPINN, a model that learns to recover Raman spectra from given CARS spectra.
  Our core methodological contribution is a physics-informed neural network that utilizes a dual-decoder architecture to disentangle resonant and non-resonant signals.
  This is done by enforcing the Kramers-Kronig causality relations via a differentiable Hilbert transform loss on the resonant and a smoothness prior on the non-resonant part of the signal.
  Trained entirely on synthetic data, RamPINN demonstrates strong zero-shot generalization to real-world experimental data, explicitly closing this gap and significantly outperforming existing baselines.
  Furthermore, we show that training with these physics-based losses alone, without access to any ground-truth Raman spectra, still yields competitive results.
  This work highlights a broader concept: formal scientific rules can act as a potent inductive bias, enabling robust, self-supervised learning in data-limited scientific domains\footnote{Project Page: \url{https://rampinn.github.io}}.
\end{abstract}

\section{INTRODUCTION}
Deep learning models require large datasets of paired inputs and ground-truth outputs to outperform existing established algorithmic approaches.
This paradigm is further exaggerated in scientific disciplines where acquiring ground-truth data is experimentally expensive.
Vibrational spectroscopy exemplifies this challenge~\parencite{sp_med,med2_ichimura2014visualizing,med3_zoladek2011non,med5_popp}.
While Raman spectroscopy provides high-fidelity molecular fingerprints~\parencite{raman1928New}, its long acquisition times limit its usage in in-situ applications~\parencite{raman_slow2023}.
A high-speed alternative, Coherent Anti-Stokes Raman Scattering (CARS), produces spectra distorted by a non-resonant background (NRB)~\parencite{cars_speed,bcars9}.
The NRB interferes coherently with the Raman signal, distorting spectral shapes and obscuring features, making its removal a challenging ill-posed inverse problem~\parencite{cars_nrb2022}.

Our approach disentangles the Raman signal from the NRB, enforcing physical constraints on each component.
We constrain the recovered Raman signal to satisfy the Kramers-Kronig (KK) relations~\parencite{kramers1928Diffusion,kronig1926dispersion,kk_book}, while modeling the NRB as a smooth component~\parencite{nrb_shape,nrb_smooth}.
These physics-based priors, embedded into the learning objective, provide sufficient supervision to recover the true Raman spectrum from a single CARS measurement.

Traditional solvers~\parencite{kk_book,interp_wp,ledht} face a circular dependency: applying the Kramers-Kronig relations requires an estimate of the non-resonant background, which itself is an unknown.
A neural network resolves this dependency by disentangling the measured spectrum into its Raman and NRB components.
The network is guided by a composite loss function that enforces the respective physical priors on each component.
This reliance on physical laws, rather than paired data, enables training exclusively on synthetically generated spectra.
We evaluate the model's zero-shot generalization on a public benchmark of six chemically diverse molecules.
This experimental setup simulates the data-scarce conditions common in this domain and validates the practical utility of our physics-informed approach.

We draw inspiration from the Physics-Informed Neural Network (PINN) paradigm~\parencite{Karniadakis2021}, but depart from its standard formulation in an essential way. 
Classical PINNs are typically developed for forward or inverse problems governed by partial differential equations, where physical knowledge enters through differential operators in the loss~\parencite{takamoto2023pdebench,moseley2021finite,HAGHIGHAT2021113741,Cai2021,JinMIO,cho2023separable,krishnapriyan2021characterizing,Maddu_2022,vemuri2024,stein2024investigating}. 
In contrast, CARS-to-Raman recovery is not naturally described by a governing differential equation. 
Instead, it is an ill-posed spectral inverse problem whose structure is determined by domain-specific physical relationships: the Kramers-Kronig (KK) relations linking the real and imaginary parts of the resonant susceptibility, and the smooth, slowly varying nature of the non-resonant background (NRB). 
We therefore extend the physics-informed learning paradigm beyond PDE-based constraints and introduce \textbf{RamPINN}, a model built around these spectroscopy-specific priors. 
Its dual-decoder architecture explicitly disentangles the resonant Raman component from the NRB, enabling us to impose distinct physical constraints on each branch: KK-consistency for the Raman component and smoothness regularization for the NRB component.

Our contributions are threefold:
\begin{enumerate*}[label=(\roman*)]
\item We propose a RamPINN: A model for solving the ill-posed inverse problem of recovering Raman spectra by embedding known physics (the Kramers-Kronig relations and smoothness prior) as a differentiable, self-supervisory loss term.
\item We demonstrate the robustness of this framework by showing it achieves state-of-the-art zero-shot generalization. Trained solely on synthetic data, our method successfully transfers to a benchmark of six diverse, real-world molecules, outperforming methods that rely on purely data-driven learning.
\item We demonstrate the efficacy of the physical priors as a supervisory signal, showing that a variant of our model trained without ground truth Raman data remains competitive with fully supervised data-driven baselines.
\end{enumerate*}
\begin{figure*}[ht]
  \centering
  \includegraphics[width=\textwidth]{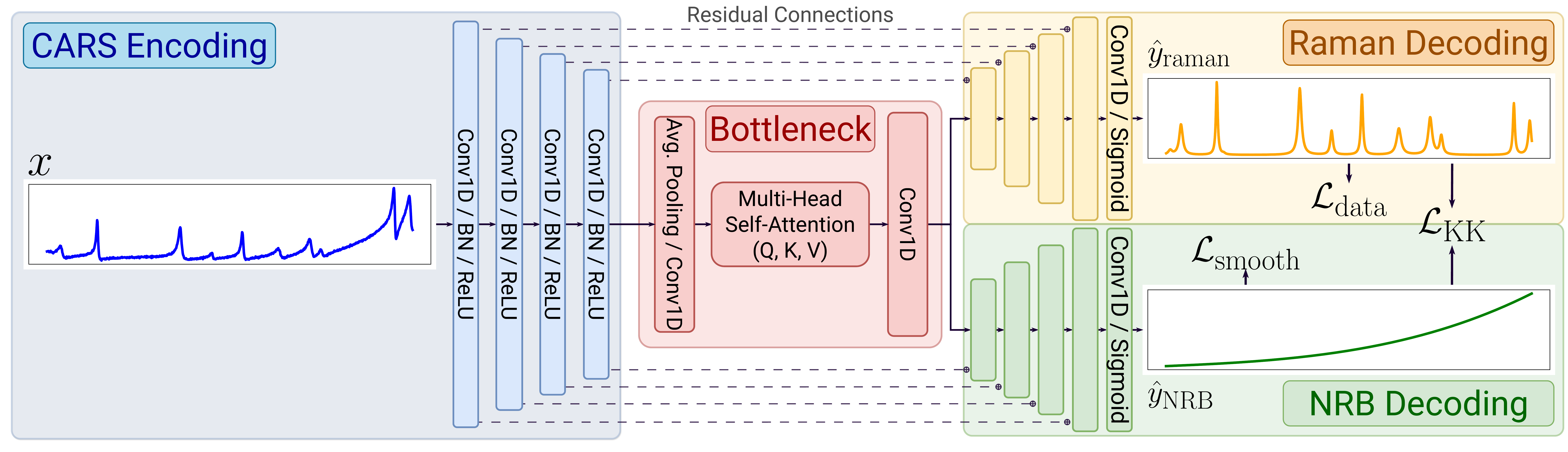}
  \caption{
    \textbf{RamPINN Architecture.}
    Our architecture builds upon recent advancements~\parencite{wang2021vector,valensise2020specnet,vaswani2017attention,ronneberger2015unet}, with a key modification: a dual-branch decoder for reconstructing Raman and non-resonant background (NRB) signals.
    We incorporate physical constraints via losses $\loss_{\mathrm{KK}}$ and $\loss_\mathrm{smooth}$ on the predicted signals.
    Note that the encoder and decoder have identical dimensions (outlined in \Cref{sec:rampinn-model-architecture}), but are depicted here with different scales for clarity.
  }
  \label{fig:rampinn-architecture}
\end{figure*}

\section{RELATED WORK}

\paragraph{Physics-informed Learning} constrains neural network training by including known physics in the loss function~\parencite{Lagaris1997}.
This idea was formalized by~\parencite{Raissi2019,Karniadakis2021,Cuomo2022}, who introduced Physics-Informed Neural Networks (PINNs), where differential equations guide the learning process.
PINNs are used to solve PDEs and ODEs~\parencite{Eivazi2022,Lin_Maxey_Li_Karniadakis_2021,vemuri2024,vemuri,cho2023separable}.
For inverse problems, PINNs learn system parameters while enforcing physical equations and are used in geophysics, structural mechanics, and other fields \parencite{soilsai,Behest,stein2024investigating,Difonzo2024}.
Beyond differential equations, domain-specific rules are integrated into the loss function for medical image registration~\parencite{wan2023warppinn} or geometry encoding~\parencite{gropp_igr, vemuri2026finr}.

\paragraph{Deep Learning for Spectroscopy} is applied to denoising~\parencite{refId0}, background correction~\parencite{Ling:85}, and Raman peak extraction~\parencite{analchem0c05391}.
Several works focused on recovering Raman spectra from corresponding CARS spectra.
SpecNet~\parencite{valensise2020specnet} introduced a CNN trained on synthetic CARS-Raman pairs to perform NRB removal, while VECTOR~\parencite{wang2021vector} used a deep autoencoder for end-to-end signal reconstruction.
Further work regularizes the process via fingerprint weights and the CH-stretching region~\parencite{vector_pinns}.
Generative approaches (GAN and CNN+GRU) were tested for NRB removal~\parencite{gan, ganrb}.
Recurrent models (LSTM~\parencite{lstm}, Bi-LSTM~\parencite{bilstm}) were explored to model spectral dependencies.
Recent approaches, such as DA-DMD, have established a decomposition-based architecture to isolate Raman peaks from NRB using known physics~\parencite{valapil2025dadmd}. 

Traditionally, works utilize physical models to embed approximations of the Kramers-Kronig (KK) relations~\parencite{kk_nrb1}, which have been improved via learned Hilbert Kernels~\parencite{ledht, mlfkk}.
In signal extraction, methods based on wavelet transform~\parencite{interp_wp, wp_mem} were also used.
However, these approaches are either limited to precomputed kernel operators or require knowledge specific to the measurement system.

In contrast, we formulate a novel physics-informed learning framework for CARS-to-Raman mapping.
We compose a loss that explicitly enforces KK-consistency via a differentiable Hilbert transform and regularizes the non-resonant background using a smoothness prior.
Therefore, the model disentangles resonant Raman features from background artifacts not only based on data but also using given rules.
This is a novel deep-learning approach for CARS spectroscopy that leverages KK physics and NRB priors within a learnable loss formulation for supervised and self-supervised training.

\section{THEORY}
We apply physics-informed learning~\parencite{Raissi2019, Karniadakis2021} to map Coherent Anti-Stokes Raman Scattering (CARS) to Raman spectra~\parencite{raman1928New}.
By embedding the problem's underlying physical laws, we improve model accuracy and robustness, particularly under data scarcity.

\subsection{Physics-Informed Learning}
Physics-informed learning is a paradigm that combines physical laws and neural networks, incorporating domain-specific knowledge into the learning process~\parencite{Raissi2019, Karniadakis2021,lagarisold}.
This paradigm trains neural networks to satisfy the underlying physical equations governing a problem task while minimizing the data-driven loss function.
These describe functional or differential relationships between input and output.

Let $x$ be the input signal, $\hat{y}$ be the prediction, and $f_\theta$ be the neural network with trainable parameters $\theta$.
A physical model expresses a constraint $\mathcal{D}$:
\begin{equation}
  \mathcal{D}(\hat{y}, x) = 0\,, \quad x \in \Omega\,,
\end{equation}
where $\mathcal{D}$ is a known operator, such as a differential or integral transform within a domain $\Omega$~\parencite{Raissi2019}.
The physics-informed loss penalizes deviations from this constraint:
\begin{equation}
  \loss_{\mathrm{phy}} = \frac{1}{|\Omega|} \int_\Omega \left| \mathcal{D}(f_\theta(x), u(x), x) \right|^2 dx~.
\end{equation}
In general this is combined with a data-driven loss term $\loss_{\mathrm{data}}$ (e.g., MSE) in a weighted objective:
\begin{equation}
  \loss_{\mathrm{total}} = \lambda_{\mathrm{data}} \cdot \loss_{\mathrm{data}} + \lambda_{\mathrm{phy}} \cdot \loss_{\mathrm{phy}}~.
\end{equation}
The physical constraint becomes part of the model's inductive bias, whose effect is controlled by the weighting parameter $\lambda_{\mathrm{phy}}$~\parencite{vemuri,mcclenny2022selfadaptive,Maddu_2022,WANG2022110768}.

\subsection{CARS-to-Raman Spectra Mapping}
This work aims to recover the Raman spectrum from a measured CARS signal.
While spontaneous Raman spectroscopy directly captures molecular vibrations~\parencite{raman1928New}, it is limited by long acquisition times~\parencite{cars_speed,cars_nrb2022}.
CARS is a faster alternative that uses nonlinear optical interactions to excite molecular vibrations more efficiently~\parencite{cars_nrb2022,cars_speed2004}.
However, the measured CARS signal is not a clean Raman readout as it contains both resonant and non-resonant components that interfere nonlinearly.
A more detailed discussion of this physical phenomenon is given in \Cref{ap:raman}.

The CARS intensity $I_{\mathrm{CARS}}(\omega)$, as a function of wavenumber $\omega$, is related to the third-order nonlinear susceptibility $\chi^{(3)}(\omega)$ by:
\begin{equation}
  I_{\mathrm{CARS}}(\omega) \propto \left| \chi^{(3)}(\omega) \right|^2 = \left| \chi_{\mathrm{R}}(\omega) + \chi_{\mathrm{NRB}} \right|^2\,,
  \label{eq:cars}
\end{equation}
where $\chi_{\mathrm{R}}(\omega)$ is the complex-valued resonant component encoding molecular vibrational information, and $\chi_{\mathrm{NRB}}$ is the real-valued non-resonant background (NRB), which is typically smooth and broad in real-world measurements~\parencite{nrb_shape,nrb_smooth}.
While the NRB amplifies signal strength, it also distorts spectral shape, making Raman reconstruction from CARS a challenging problem.

To constrain this ill-posed problem, we leverage two well-established physical priors.
\paragraph{The Kramers-Kroning relation}

The resonant susceptibility $\chi_{\mathrm{R}}$ satisfies the Kramers-Kronig (KK) relation~\parencite{kk_book,kramers1928Diffusion,kronig1926dispersion}, which links its real and imaginary parts through causality in linear response theory~\parencite{wang2021vector,bilstm}. This is further explained in \Cref{ap:kk}.
Specifically,
\begin{equation}
  \chi_{\mathrm{R}} (\omega) = \Re[\chi_{\mathrm{R}}(\omega)] \pm \Im[(\chi_{\mathrm{R}}(\omega))]
\end{equation}
and
\begin{equation}
  \Re[\chi_{\mathrm{R}}(\omega)] = \mathcal{H}(\Im[\chi_{\mathrm{R}}(\omega)])\,,
\end{equation}
where \( \mathcal{H}(\cdot) \) is the Hilbert transform~\parencite{Graf2010}.
The imaginary part carries the true Raman spectrum~\parencite{muller2007Coherent}.
This relation provides a consistency constraint between what we want to recover ($\Im(\chi_{\mathrm{R}})$) and what we estimate from the observed signal (the real part $\Re(\chi_{\mathrm{R}})$).

To derive the specific loss function we use, we expand \Cref{eq:cars} term (following \cite{valensise2020specnet}), 
\begin{equation}
    \left| \chi_{\mathrm{NRB}} + \chi_{\mathrm{res}} \right|^{2}
= |\chi_{\mathrm{NRB}}|^{2} + |\chi_{\mathrm{res}}|^{2} +
 2\,\Re\!\big(\chi_{\mathrm{NRB}}\chi_{\mathrm{res}}\big)
\end{equation}

Here, we assume the resonant component is weak compared to the NRB $|\chi_{\mathrm{res}}| \ll |\chi_{\mathrm{NRB}}|$ ( \cite{valensise2020specnet}), such that we can neglect the purely resonant term $|\chi_{\mathrm{res}}|^{2}$ and obtain
\begin{equation}
I_{\mathrm{CARS}} \approx |\chi_{\mathrm{NRB}}|^{2} + 2\,\Re\!\big(\chi_{\mathrm{NRB}}\chi_{\mathrm{res}}\big).
\label{eq:i_cars_approx}
\end{equation}

The non-resonant background is considered real-valued (instantaneous electronic response)
\begin{equation}
\chi_{\mathrm{NRB}} \in \mathbb{R},
\end{equation}

So the term in the RHS of \ref{eq:i_cars_approx} simplifies to
\begin{equation}
\Re\!\big(\chi_{\mathrm{NRB}}\chi_{\mathrm{res}}\big) = \chi_{\mathrm{NRB}}\,\Re[\chi_{\mathrm{res}}].
\end{equation}
Substituting this term into the \ref{eq:i_cars_approx} results in
\begin{equation}
I_{\mathrm{CARS}} \approx \chi_{\mathrm{NRB}}^{2} + 2\,\chi_{\mathrm{NRB}}\,\Re[\chi_{\mathrm{res}}].
\end{equation} 

As is standard in KK-based CARS processing (e.g., \cite{ledht}, \cite{kk_nrb1}), one works with a normalized CARS signal obtained by dividing out the NRB envelope (the instrument is assumed to produce normalized spectra). We define:
\begin{equation}
x := \frac{I_{\mathrm{CARS}}}{\chi_{\mathrm{NRB}}},
\end{equation}
which yields
\begin{equation}
x \approx \chi_{\mathrm{NRB}} + 2\,\Re[\chi_{\mathrm{res}}].
\end{equation}

Raman spectra are interpreted on an overall scale (peak positions and shapes are important, not absolute magnitudes).
The KK relation constrains the functional dependence of real and imaginary parts, not their absolute scale. For this reason, we use the KK-consistent imaginary component, :
\begin{equation}
\Im\!\left(\mathcal{H}(x - \hat{y}_{\mathrm{NRB}})\right)
\end{equation}
as the physics-based target for the Raman branch.

\paragraph{NRB characteristics}
The background in our signal or NRB is smooth and lacks sharp peaks in real-world measurements~\parencite{nrb_shape,nrb_smooth}.
This motivates a regularization term penalizing high curvature in the estimated background.
These constraints directly integrate physical principles into the learning objective: one loss term promotes Kramers-Kronig (KK) consistency in the predicted Raman spectrum, while the other enforces smoothness in the NRB estimate~\parencite{bilstm, gan}.
Consequently, our approach is both data-driven and physics-grounded.
This integration of physical priors fundamentally distinguishes RamPINN from previous, purely data-driven methods.

\section{RamPINN - METHODOLOGY}
A CARS spectrum is represented as a function  $x(\omega)$, where $\omega$ denotes the wavenumber, and the function value represents the measured relative intensity at that wavenumber.
Similarly, the corresponding Raman spectrum and non-resonant background (NRB) are given by $y_{\mathrm{raman}}(\omega)$ and $y_{\mathrm{NRB}}(\omega)$, respectively.

The wavenumber ($\omega$) is fixed across all samples during training and inference. We simplify notation by dropping the explicit dependence on $\omega$, and refer to the spectra as vectors: $x$, $y_{\mathrm{raman}}$, and $y_{\mathrm{NRB}}$.
All spectra are normalized, as we are concerned with relative spectral shapes (signatures) rather than absolute intensity values.

We aim to learn a mapping from the CARS spectrum to its underlying Raman and NRB components.
A neural network $f_\theta$, parameterized by $\theta$, takes the normalized CARS spectrum $x$ as input and reconstructs the Raman ($\hy_\mathrm{raman}$) and NRB ($\hy_\mathrm{NRB}$) spectra tuple:

\begin{equation}
  f_\theta(x) = \left( \hy_{\mathrm{raman}}, \hy_{\mathrm{NRB}} \right)~.
\end{equation}

We aim to design a learning framework that enforces this decomposition in a way that respects the known physics of the CARS process.
In the following sections, we describe how we achieve this using physics-informed loss functions based on the Kramers-Kronig relations and the smooth, peak-free nature of the NRB.

\subsection{Model Architecture}
Physics-informed learning is architecture-agnostic, yet recent advancements motivate our components~\parencite{valensise2020specnet,wang2021vector,vaswani2017attention,ronneberger2015unet}.
We use a 1D Convolutional U-Net~\parencite{ronneberger2015unet} tailored for spectral signal decomposition, shown in \Cref{fig:rampinn-architecture}.
The model takes raw CARS spectra as input and outputs two components: the resonant Raman signal and the non-resonant background (NRB).
The network employs a shared encoder and a dual-branch decoder for disentanglement, which are trained jointly under the physics-informed objective in \Cref{eq:loss_total}.

The encoder consists of four convolutional blocks with average pooling to preserve smooth signal features, progressively reducing temporal resolution while increasing feature depth.
A self-attention block~\parencite{vaswani2017attention} captures long-range dependencies in the spectral domain, necessary for resolving overlapping peaks.

A dual-branch decoder reconstructs the Raman signal and NRB.
Each branch comprises four upsampling blocks, a final convolution, and a sigmoid activation.
To avoid checkerboard artifacts, we use upsampling followed by a one-dimensional convolutional layer rather than transposed convolutions~\parencite{hock2022n2v2,buchner2023improved,buchner2025electromyography,kwarciak2024unsupervised}.
Skip connections at each level concatenate encoder features with the corresponding decoder inputs, helping preserve fine-grained spectral structure throughout reconstruction.

Overall, this architecture is designed to decompose the input spectrum into physically meaningful components while maintaining flexibility for both supervised and physics-guided training.
In \Cref{sec:rampinn-model-architecture}, we detail each component of the RamPINN model architecture.
We also discuss the selection of our chosen backbone, comprising a U-Net~\parencite{ronneberger2015unet} with self-attention~\parencite{vaswani2017attention}, over other existing backbone architectures.

\subsection{Optimization}
Let $x$ be the measured CARS spectrum, and the goal is to predict $\hy = (\hy_{\mathrm{raman}}, \hy_{\mathrm{NRB}})$, the Raman-resonant and non-resonant components, respectively, using a neural network $f_\theta(x)$.
The Kramers-Kronig (KK) relations govern the underlying physical relationship~\parencite{kk_book,kramers1928Diffusion,kronig1926dispersion,guenther2004Encyclopedia}.
A detailed description of Raman spectroscopy and Kramers-Kronig relationships is provided in \Cref{ap:raman,ap:kk}, respectively.

\paragraph{Kramers-Kronig Regularization.}
Causality in optical response implies that the Raman component must be consistent with the imaginary part of the Hilbert transform of the residual signal~\parencite{kk_nrb1,kk_book}.
After subtracting the NRB, the residual signal should follow:
\begin{equation}
  \loss_{\mathrm{KK}} = \lvert \hy_{\mathrm{raman}} - \Im \left( \mathcal{H}(x - \hy_{\mathrm{NRB}}) \right) \rvert^2~.
\end{equation}

Here, $\mathcal{H}(\cdot)$ denotes the differentiable Hilbert transform, which produces the analytic signal whose imaginary part corresponds to the KK-paired component.
A complete formulation is provided in \Cref{sec:kramers--kronig-implementation}.

\paragraph{NRB Regularization.}
The NRB is smooth and broad~\parencite{nrb_smooth,wang2021vector}.
We enforce this behavior by regularizing the derivative of the predicted NRB signal, penalizing rapid changes~\parencite{spectralpinn, pmlr-v137-rosca20a}:
\begin{equation}
  \loss_{\mathrm{smooth}} = \lvert \nabla \hy_{\mathrm{NRB}} \rvert^2~.
\end{equation}

\paragraph{Optimization Term.}
The loss is a weighted sum of a data fidelity term and the physics-based regularizers:
\begin{equation}
  \loss_{\mathrm{total}} =
  \lambda_{\mathrm{data}}  \loss_{\mathrm{data}} +
  \lambda_{\mathrm{KK}}     \loss_{\mathrm{KK}} +
  \lambda_{\mathrm{smooth}} \loss_{\mathrm{smooth}}\,.
  \label{eq:loss_total}
\end{equation}

\begin{figure*}[ht!]
    \centering
    \begin{subfigure}[t]{0.335\textwidth}
        \centering
        \includegraphics[width=\textwidth]{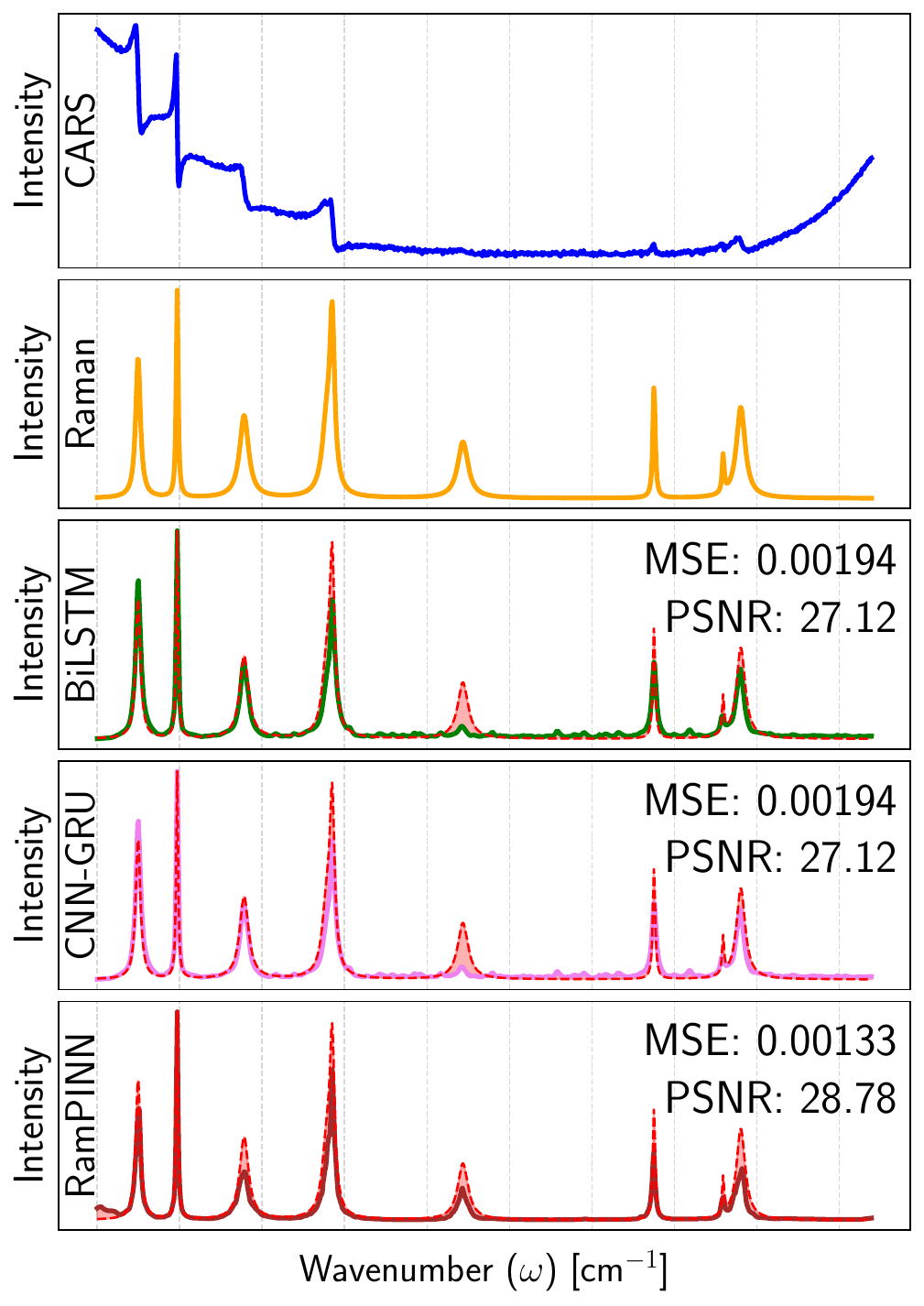}
        \caption{}
        \label{fig:top-3-1}
    \end{subfigure}
    \hfill
    \begin{subfigure}[t]{0.319\textwidth}
        \centering
        \includegraphics[width=\textwidth]{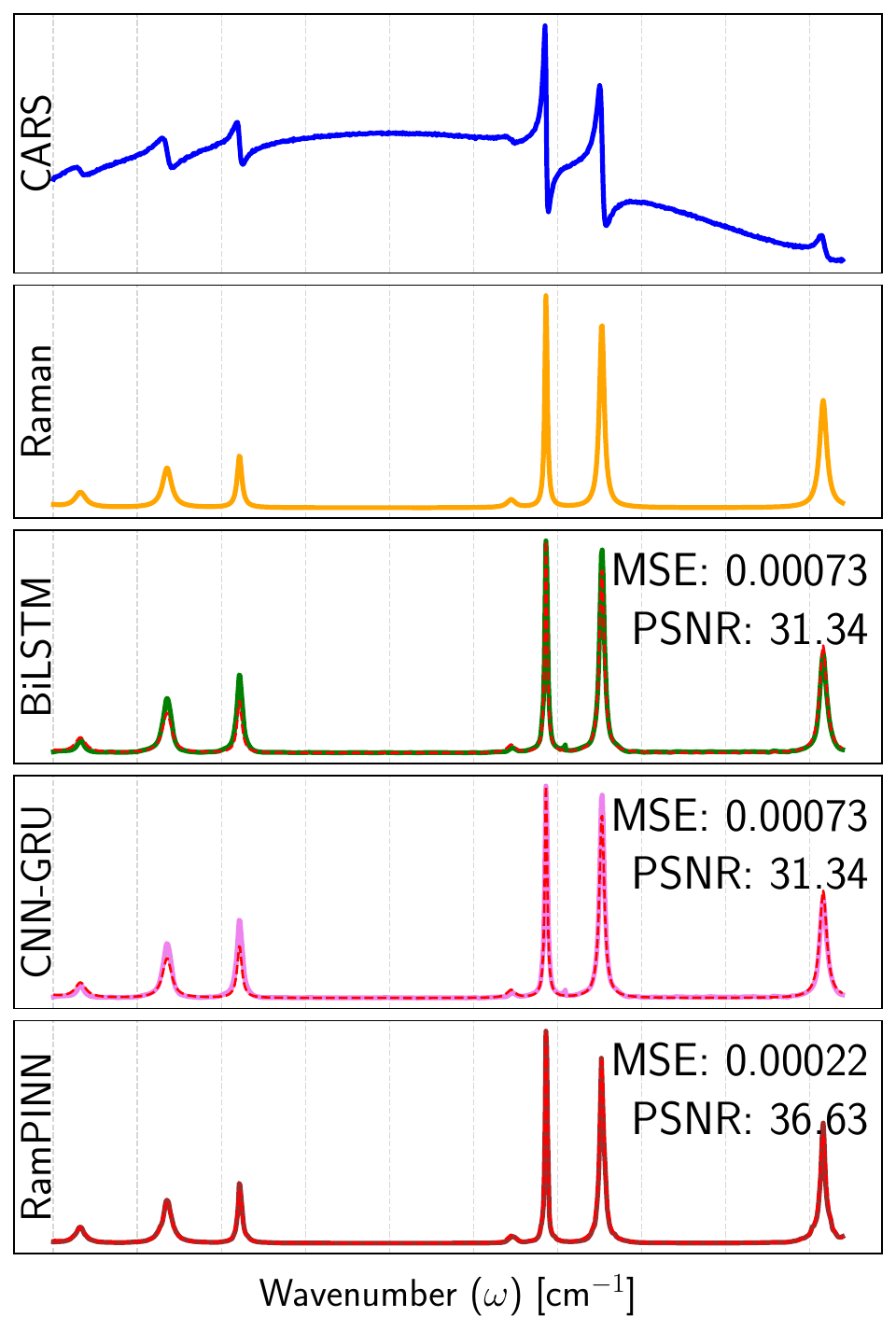}
        \caption{}
        \label{fig:top-3-2}
    \end{subfigure}
    \hfill
    \begin{subfigure}[t]{0.319\textwidth}
        \centering
        \includegraphics[width=\textwidth]{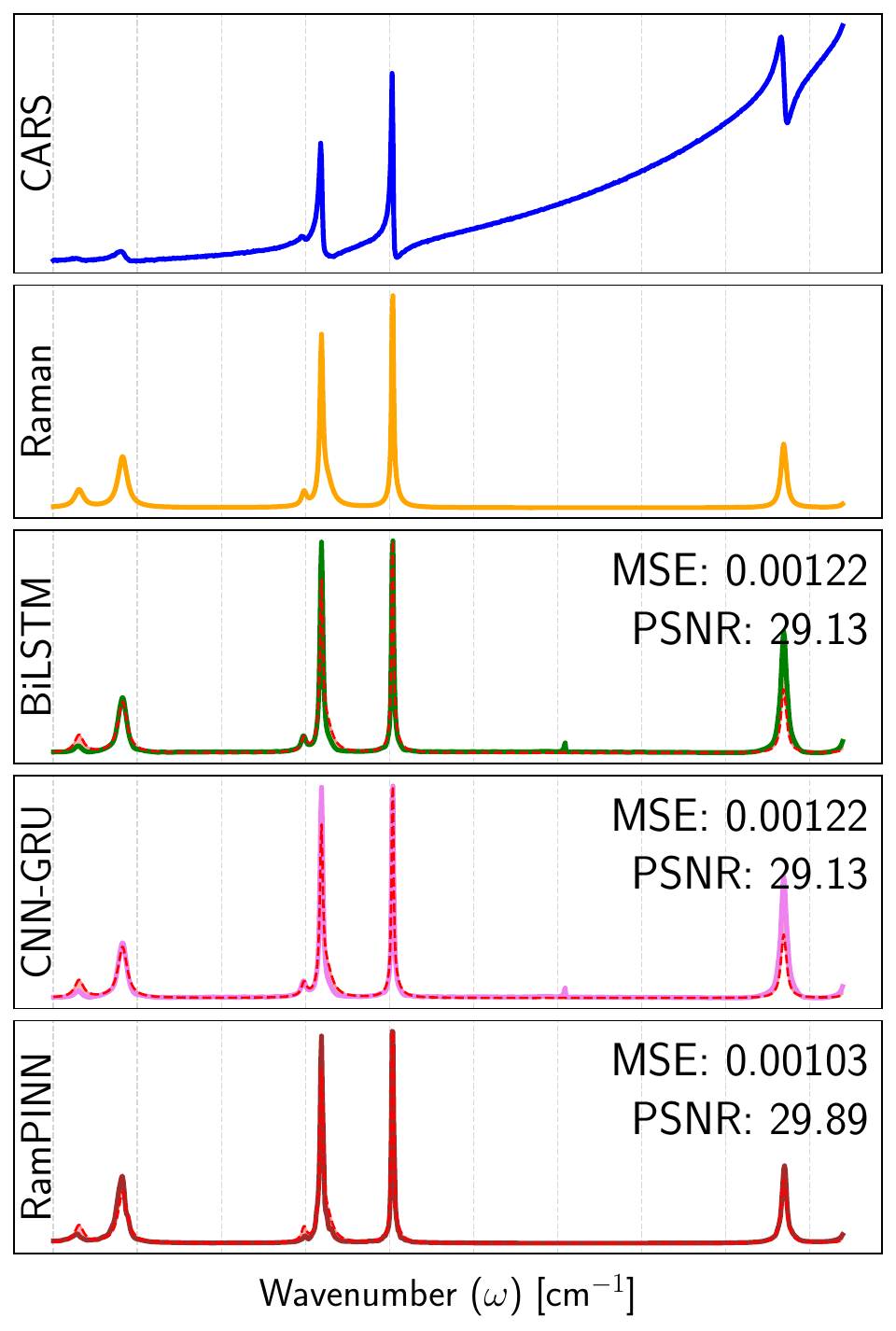}
        \caption{}
        \label{fig:top-3-3}
    \end{subfigure}
    \caption{
        \textbf{Qualitative Comparison of Raman Signal Extraction.}
        We visualize the Raman reconstruction of RamPINN, BiLSTM, and CNN-GRU on three synthetic samples (\subref{fig:top-3-1}-\subref{fig:top-3-3}).
        RamPINN outperforms the other methods, both qualitatively and quantitatively, as shown by the error values and lines (best viewed digitally).
        We provide additional plots in \Cref{ap:reconstruction-plots}.
    }
    \label{fig:qualitative-raman-reconstruction}
\end{figure*}

\subsection{Implementation}
We train RamPINN to learn the mapping from CARS spectra to their corresponding Raman and NRB components.
The training algorithm, including the differentiable Hilbert transform for computing the Kramers-Kronig loss, is outlined in \Cref{sec:kramers--kronig-implementation}.

Otherwise, we use standard deep learning training practices - Adam optimizer~\parencite{kingma2014adam} with a learning rate of $10^{-3}$ and early stopping based on validation loss.
All models are implemented in PyTorch~\parencite{paszke2019pytorch} with mini-batch training, gradient clipping, and loss normalization where appropriate.
We fix the values of the regularization terms (obtained by an initial hyperparameter search) for RamPINN:
$\lambda_{\mathrm{smooth}} = 10$, $\lambda_{\mathrm{data}} = 10$, and $\lambda_{\mathrm{KK}} = 1$.

\section{EXPERIMENTS AND RESULTS}
We examine whether including physical knowledge during training enhances performance compared to data-driven baselines.
Therefore, we compare RamPINN, supervised and self-supervised (by setting $\lambda_\mathrm{data}=0$),  against state-of-the-art deep learning CARS-to-Raman approaches: SpecNet~\parencite{valensise2020specnet}, VECTOR~\parencite{wang2021vector}, LSTM~\parencite{lstm}, BiLSTM~\parencite{bilstm}, GAN~\parencite{gan}, and CNN-GRU~\parencite{gan}.
We follow their recommended training procedures.
More details for these models are provided in \Cref{sec:baselines}.
All models have a comparable capacity, ensuring that any observed improvement can be attributed to the training methodology, specifically, the integration of physics-based constraints, rather than differences in model size.
We also include traditional non-deep learning approaches for comparison:  TDKK~\parencite{kk_nrb1}, LeDHT~\parencite{ledht}, and IWT~\parencite{interp_wp}.

Due to the limited availability of real paired CARS and Raman spectra, we focus on synthetic training data.
Our synthetic dataset mimics experimental conditions, described in \Cref{sec:dg}.
We use 2000 synthetic samples, 1000 for training and 1000 for testing.
All models are trained and evaluated (with an NVIDIA GTX 1080) on this common dataset and further tested on six public real-world examples for zero-shot signal extraction performance evaluation~\parencite{real_sample}.

We evaluate performance using mean squared error (MSE) and Peak Signal-to-Noise ratio (PSNR (dB)) between predicted and ground truth spectra.
To ensure robustness, we report results over 10 independent training runs.
Quantitative results are given in \Cref{tab:experiment}.
Beyond MSE/PSNR, we also report peak-aware metrics that assess whether the predicted spectra recover the correct peaks, how well their positions align, and how accurately their magnitudes are reproduced.
Complete definitions and results (F1 for peak detection, normalized mean location error, and relative intensity error) are provided in \Cref{app:metrics}.

The supervised RamPINN outperforms all baselines by a large margin, achieving the lowest MSE and highest PSNR on the test set.
This demonstrates the benefit of incorporating known physics into the learning process.
Notably, the self-supervised RamPINN variant ($\lambda_\mathrm{data}=0$) remains competitive, outperforming most data-driven baselines that rely on ground truth data.
This supports our hypothesis that leveraging physical knowledge improves performance, even with limited or no reconstruction regularization.
This demonstrates that physical priors can effectively replace experimentally expensive paired data, enabling high-fidelity spectral recovery in data-limited settings.

The qualitative results for the top three performing models (RamPINN (supervised), BiLSTM~\parencite{bilstm}, and CNN-GRU~\parencite{gan}), as shown in \Cref{fig:qualitative-raman-reconstruction}, further support these findings.
Visually, RamPINN accurately reconstructs the Raman spectra from three synthetic CARS, especially recovering smaller peaks that other models miss.
The \Cref{ap:reconstruction-plots} provides more examples.
Although synthetic results are encouraging, zero-shot performance on real-world data is crucial for gauging practical applicability.

\begin{table}[t]
  \centering
  \caption{
    \textbf{Quantitative Comparison of Raman Reconstruction Methods.}
    MSE and PSNR are computed between the predicted and ground truth Raman spectra on the test set.
    Lower MSE and higher PSNR indicate better performance.
    The top three models are ranked gold\gm, silver\sm, and bronze\bm, respectively, based on MSE values (when two MSEs are the same, we use PSNR).
    All deep learning results are given over 10 independent runs.
    Traditional methods (denoted with \textdagger) estimate a single deterministic spectrum for comparison and require an NRB estimate.
  }
  \begin{adjustbox}{max width=\linewidth}
\begin{tabular}{lp{0.01cm}lp{0.05cm}l}
  \toprule
  \textbf{Method} & & \multicolumn{1}{c}{\textbf{MSE} $\downarrow$} & & \multicolumn{1}{c}{\textbf{PSNR (dB)} $\uparrow$}\\
  \midrule
  TDKK\textsuperscript{$\dagger$}                 &   & 0.0283                              &  & 15.48 \\
  LeDHT\textsuperscript{$\dagger$}                  &   & 0.0814                              &  & 10.91\\
  IWT\textsuperscript{$\dagger$}                &   & 0.0139                              &  & 18.75\\
  \midrule
  SpecNet &   & 0.0064 {\footnotesize$\pm$ 0.0003}  &   & 21.91 {\footnotesize$\pm$ 0.21} \\
  VECTOR       &   & 0.0205 {\footnotesize$\pm$ 0.0002}  &   & 16.88 {\footnotesize$\pm$ 0.04} \\
  LSTM                    &   & 0.0732 {\footnotesize$\pm$ 0.1169}  &   & 19.30 {\footnotesize$\pm$ 8.77} \\
  BiLSTM               &\sm& 0.0007 {\footnotesize$\pm$ 0.0002}  &\sm& 31.57 {\footnotesize$\pm$ 1.29} \\
  GAN                     &   & 0.0088 {\footnotesize$\pm$ 0.0118}  &   & 22.74 {\footnotesize$\pm$ 3.74} \\
  CNN-GRU                  &\bm& 0.0019 {\footnotesize$\pm$ 0.0004}  &\bm& 27.38 {\footnotesize$\pm$ 0.97} \\
  \midrule
  RamPINN                             &\gm& 0.0006 {\footnotesize$\pm$ 0.0001}  &\gm& 33.83 {\footnotesize$\pm$ 0.13} \\
  RamPINN (Self-sup)                  &   & 0.0053 {\footnotesize$\pm$ 0.0003}  &   & 22.79 {\footnotesize$\pm$ 0.28} \\
  \bottomrule
\end{tabular}
\end{adjustbox}

  \label{tab:experiment}
\end{table}

\subsection{Zero-shot Evaluation on Real-world Samples}

To evaluate how well our models generalize beyond synthetic data, we conduct a zero-shot test: models trained entirely on synthetic spectra are applied directly to real CARS measurements, without any fine-tuning or adaptation.
We use six publicly available samples, Acetone, DMSO, Ethanol, Isopropanol, Methanol, and Toluene with paired CARS and Raman spectra from~\parencite{real_sample,realdata_zenodo}.

\begin{figure}[t]
  \centering
  \includegraphics[width=\linewidth]{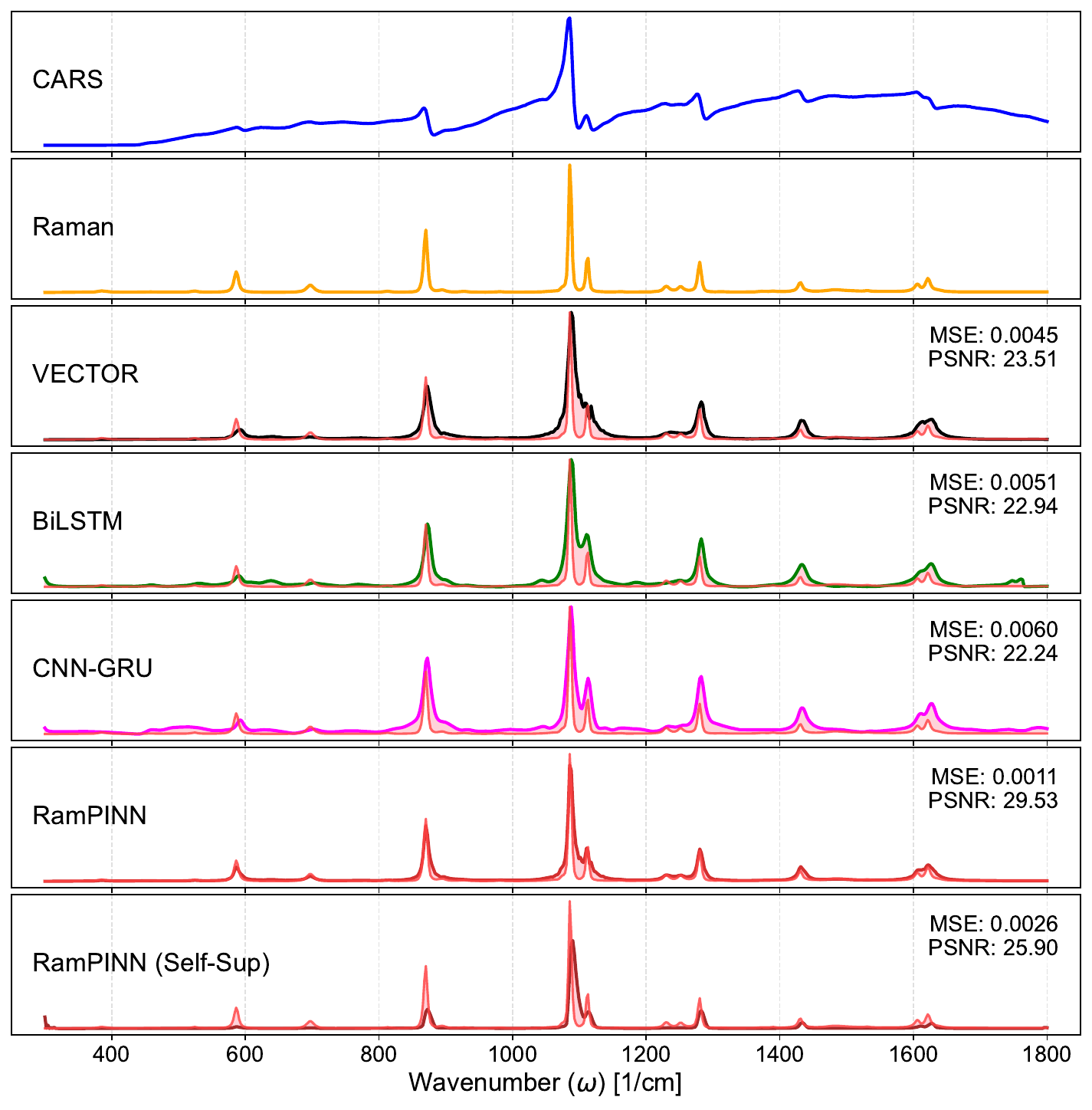}
  \caption{
    \textbf{Zero-shot Raman Spectra Extraction -- Toluene.}
    Looking at input CARS, ground truth Raman, and reconstructions from baseline methods (VECTOR, BiLSTM, CNN-GRU), our RamPINN approach provides a more accurate zero-shot Raman spectra recovery, achieving the lowest MSE and highest PSNR.
  }
  \label{fig:realworld-toluene}
\end{figure}

We compare RamPINN predictions (supervised and self-supervised variants) with other baseline deep learning models against the ground-truth Raman spectra.
For each model, we report the best-performing version.
The goal is to assess how well each model handles real experimental signals without domain-specific calibration.
Results are summarized in \Cref{tab:experiment-realworld}, with one representative prediction of Toluene shown in \Cref{fig:realworld-toluene}.
We can observe here that RamPINN reconstructs all peaks with accurate shape and intensity at correct locations.
This is also seen for other samples for which qualitative visualizations are included in \Cref{ap:reconstruction-plots}.

These results indicate that sufficient synthetic data with self-supervision translates well to real-world applications.
RamPINN consistently outperforms all purely data-driven baselines, demonstrating that embedding physical constraints leads to better generalization on real-world data.
Notably, even the self-supervised RamPINN, trained without ground truth spectra, still generates competitive reconstructions, rivaling supervised baselines.
This highlights the strength of physics-informed training, especially in settings where ground truth data are scarce or unavailable.

\begin{table*}[t]
  \centering
  \caption{
    \textbf{Real-world Zero-shot Evaluation.}
    Zero-shot evaluation results on six real-world CARS samples (Acetone, DMSO, Ethanol, Isopropanol, Methanol, Toluene)~\parencite{real_sample,realdata_zenodo}.
    RamPINN shows strong performance, consistently achieving the lowest Mean Squared Error (MSE) and highest Peak Signal-to-Noise Ratio (PSNR) across all samples, outperforming the baseline methods.
    The self-supervised RamPINN variant (Self-sup) also notably surpasses many baselines.
    The top three models are ranked gold\gm, silver\sm, and bronze\bm, respectively, based on MSE values (when two MSEs are the same, we use PSNR).
  }
  \label{tab:experiment-realworld}
  \resizebox{\textwidth}{!}{%
	\begin{tabular}{l rr rr rr rr rr rr}
		\toprule
		& \multicolumn{2}{c}{Acetone} & \multicolumn{2}{c}{DMSO} & \multicolumn{2}{c}{Ethanol} & \multicolumn{2}{c}{Isopropanol} & \multicolumn{2}{c}{Methanol} & \multicolumn{2}{c}{Toluene}                                          \\
		\cmidrule(lr){2-3}\cmidrule(lr){4-5}\cmidrule(lr){6-7}\cmidrule(lr){8-9}\cmidrule(lr){10-11}\cmidrule{12-13}
		                                    & MSE $\downarrow$    & PSNR $\uparrow$ & MSE $\downarrow$    & PSNR $\uparrow$ & MSE $\downarrow$    & PSNR $\uparrow$ & MSE $\downarrow$    & PSNR $\uparrow$ & MSE $\downarrow$    & PSNR $\uparrow$  & MSE $\downarrow$    & PSNR $\uparrow$ \\
		\midrule
		TDKK                 & 0.0283              & 15.48           & 0.0098              & 20.04           & 0.0140              & 18.30           & 0.0192              & 17.15           & 0.0064              & 22.06            & 0.0064              & 31.92           \\
		LeDHT                  & 0.2663              & 5.74            & 0.0157              & 18.01           & 0.0351              & 14.53           & 0.0607              & 12.16           & 0.0315              & 15.01            & 0.1129              & 9.47            \\
		IWT                & 0.0078              & 21.02           & 0.0083              & 20.07           & 0.0149              & 18.26           & 0.0182              & 17.39           & \sm 0.0031          & 24.97            & 0.0061              & 22.97           \\
		\midrule
		SpecNet & 0.0108              & 19.65           & 0.0046              & 23.42           & 0.0036              & 24.43           & 0.0169              & 17.73           & 0.0041              & 23.88            & 0.0061              & 22.12           \\
		VECTOR        & \bm 0.0041          & 23.88           & 0.0051              & 22.96           & \bm 0.0033          & 24.84           & 0.0105              & 19.78           & 0.0039              & 24.12            & 0.0045              & 23.51           \\
		LSTM                    & 0.0122              & 19.12           & 0.0101              & 19.95           & 0.0131              & 18.83           & 0.0165              & 17.82           & 0.0555              & 12.56            & 0.0075              & 21.25           \\
		BiLSTM                & 0.0043              & 23.67           & 0.0046              & 23.39           & 0.0041              & 23.92           & \bm 0.0100          & 20.00           & 0.0037              & 24.30            & 0.0051              & 22.94           \\
		GAN                      & \sm 0.0035          & 24.53           & \bm 0.0044          & 23.57           & 0.0046              & 23.38           & 0.0153              & 18.16           & 0.0037              & 24.27            & \bm 0.0031          & 25.05           \\
		CNN-GRU                  & 0.0105              & 19.80           & \sm 0.0044          & 23.58           & \sm 0.0033          & 24.85           & 0.0169              & 17.72           & \bm 0.0037          & 24.32            & 0.0060              & 22.24           \\
		\midrule
		RamPINN                             & \gm\textBF{0.0011} & \textBF{29.69}  & \gm \textBF{0.0013} & \textBF{28.98}  & \gm \textBF{0.0008} & \textBF{30.87}  & \gm \textBF{0.0026} & \textBF{25.80}  & \gm \textBF{0.0010} & \textBF{30.14}   & \gm \textBF{0.0011} & \textBF{29.53}  \\
		RamPINN (Self-sup)                  & 0.0053              & 22.79           & 0.0076              & 21.20           & 0.0099              & 20.02           & \sm 0.0035            & 24.56           & 0.0114              & 19.43            & \sm 0.0026          & 25.90           \\
		\bottomrule
	\end{tabular}%
}

\end{table*}

\section{ABLATION STUDIES}
We conduct ablations to assess the key parts of our method.
Specifically, we: (1)Quantify the impact of our physics-informed KK loss and smoothness prior. (2)Analyze performance in relation to the amount of training data. (3)Test the model's robustness to violations of the non-resonant background (NRB) smoothness assumption.
These studies probe the method's limitations and validate its suitability for practical application.

\subsection{Effect of the Physics Loss Term}
One of our contributions is the incorporation of the Kramers-Kronig (KK) loss ($\loss_{\mathrm{KK}}$), which isolates and quantifies its effect compared to purely data-driven approaches.
Specifically, we vary the weight of the physics-based KK loss term, denoted by $\lambda_{\mathrm{KK}} $, from zero to one in ten equidistant intervals.

As shown in \Cref{fig:ablation-a}, when $\lambda_{\mathrm{KK}}=0$, the model reduces to a data-driven approach without any physics constraint, and it shows weaker performance.
As the value of $\lambda_{\mathrm{KK}}$ increases, the influence of physics during training increases, leading to better performance.
Hence, we quantify how much the KK loss $\loss_\mathrm{KK}$ contributes to model performance.

\subsection{Effect of Data}
\begin{figure*}[htbp!]
  \centering
  \begin{subfigure}[b]{0.30\textwidth}
    \centering        \includegraphics[width=\linewidth]{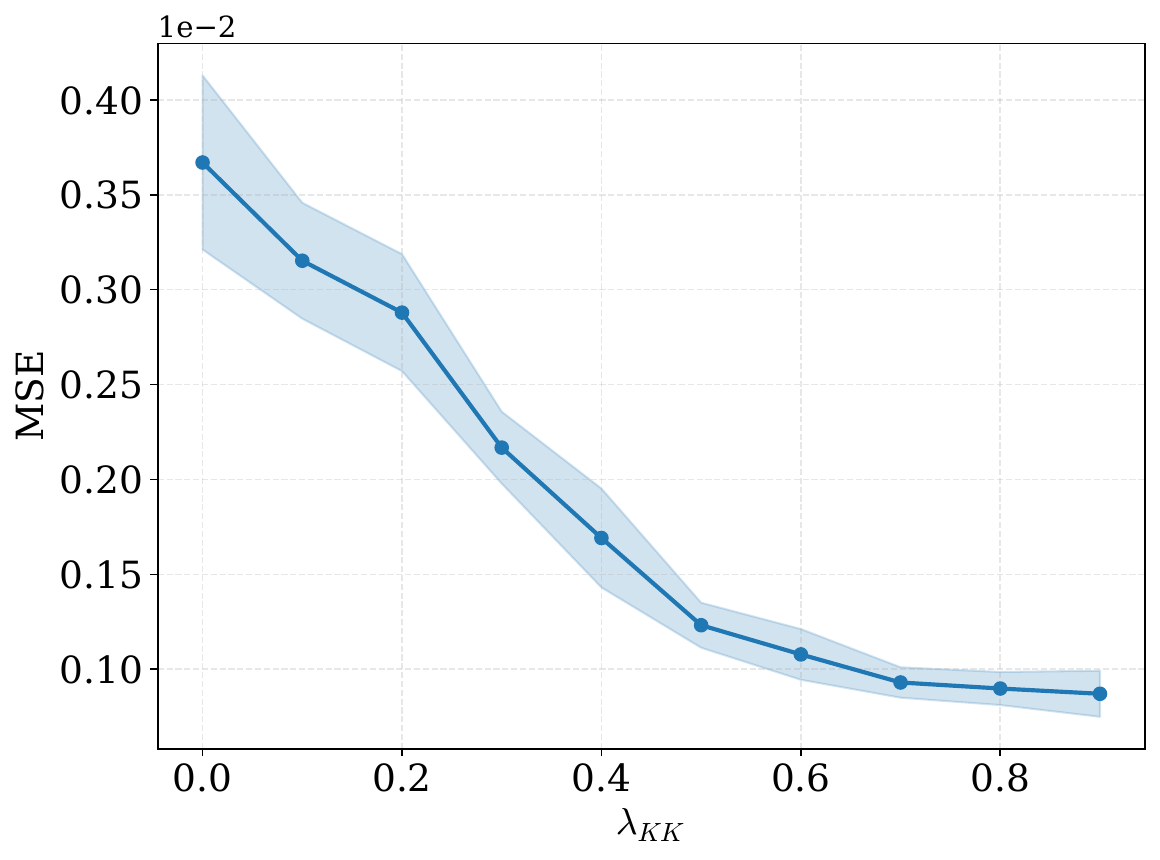}
    \caption{
      \textbf{Effect of $\loss_{\mathrm{KK}}$}
    }
    \label{fig:ablation-a}
  \end{subfigure}
  \hfill
  \begin{subfigure}[b]{0.30\textwidth}
    \centering
    \includegraphics[width=\linewidth]{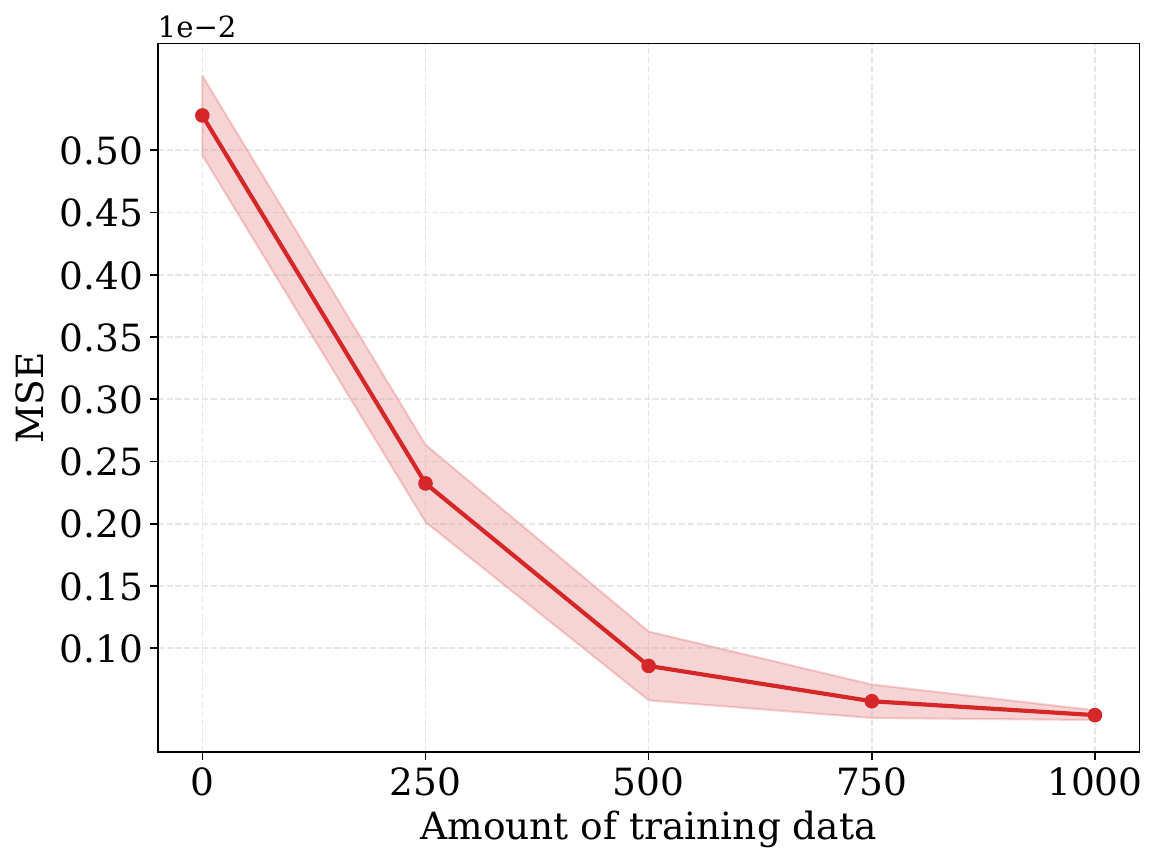}
    \caption{
      \textbf{Effect of $\loss_{\mathrm{data}}$}
    }
    \label{fig:ablation-b}
  \end{subfigure}
  \hfill
  \begin{subfigure}[b]{0.30\textwidth}
    \centering
    \includegraphics[width=\linewidth]{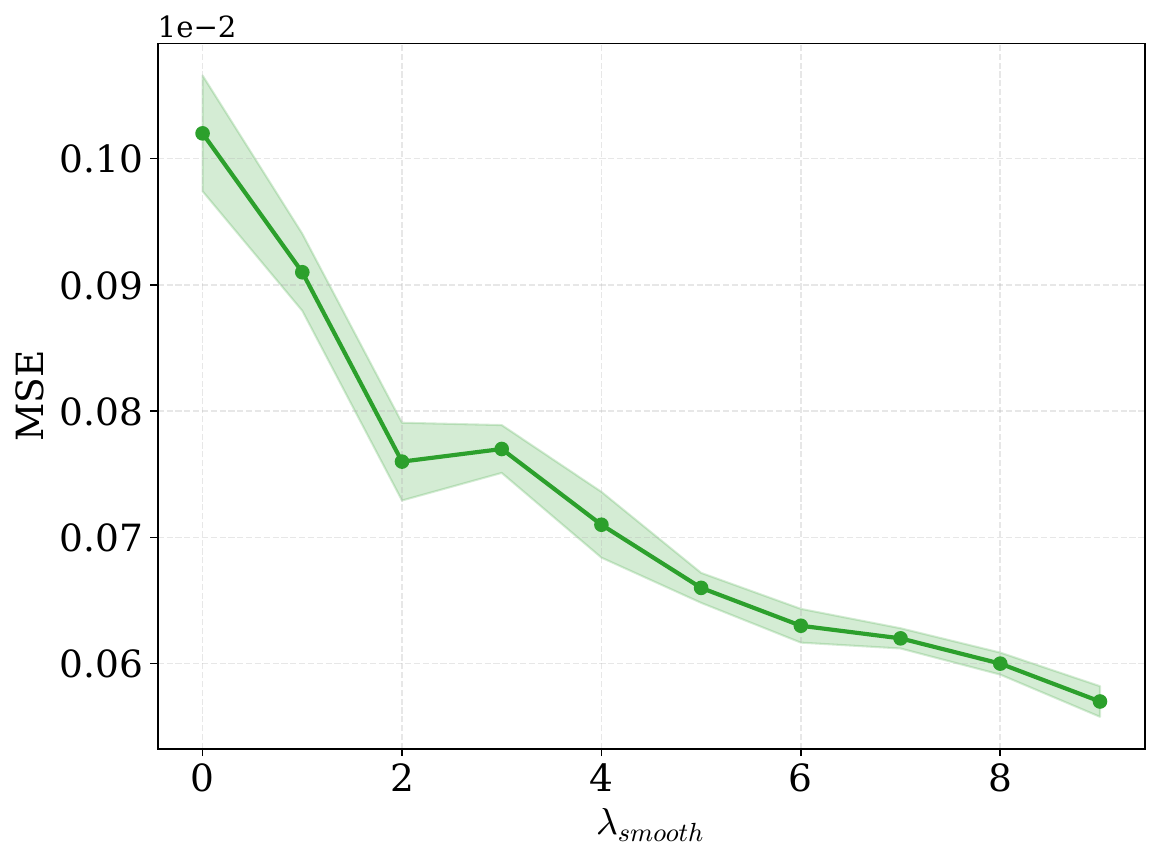}
    \caption{
      \textbf{Effect of $\loss_{\mathrm{smooth}}$}
    }
    \label{fig:ablation-c}
  \end{subfigure}
  \caption{
    \textbf{Ablation Study Quantifying the Effects of Individual Loss Terms on Test MSE}.
    We observe that scaling our physics-based Kramers-Kronig regularization term, $\loss_{\mathrm{KK}}$, steadily decreases the test MSE with increasing influence.
    We also see similar effects with increasing the amount of training data ($\loss_{\mathrm{Data}}$) and the smoothness constraint imposed on NRB ($\loss_{\mathrm{smooth}}$).
  }
  \label{fig:ablations}
\end{figure*}

While we show that the Kramers-Kronig (KK) loss enhances standard data fidelity terms, it is crucial to test scaling the training data.
We freeze the test set and vary the number of training samples from $0$ to $1000$, keeping all other settings fixed.
This setup ensures systematic evaluation of the impact of supervision concerning generalization.
Performance improves with more training data, as shown in \Cref{fig:ablation-b}.
However, even in the fully self-supervised case, where RamPINN is trained without any data signal, the model recovers meaningful Raman spectra from CARS inputs.
While not matching the accuracy of supervised models, these self-supervised predictions remain competitive, underscoring the strength of the physics-informed loss as an inductive bias.
We observe diminishing returns with more data, suggesting that either the synthetic data lacks diversity or that the model has reached its representational capacity.
The results show that the problem can be handled effectively with limited supervision.

\subsection{Effect of Smoothness}
We ablated the influence of $\lambda_\mathrm{smooth}$.
This prior is crucial for disentangling the signal, as it prevents the NRB decoder from fitting sharp Raman peaks to the synthetic data.
While the KK-loss has a larger impact, the smoothness prior consistently improves performance as shown in \Cref{fig:ablation-c}.

\subsection{Additional Robustness Studies}
A key question is how robust RamPINN is to real-world experimental complexities, such as nonlinearities.
To test these failure modes, we add artifacts.
We conducted robustness studies, presented in full in \Cref{ap:rob}.
We evaluated the model's performance against increasingly nonlinear NRB shapes and in the presence of sharp, peak-like artifacts designed to simulate instrumental noise.
Our results consistently show that the model exhibits graceful degradation rather than catastrophic failure.
RamPINN effectively learns to attribute non-physical artifacts to the NRB component, preserving the fidelity of the Raman reconstruction even under significant violations of assumptions.

\section{LIMITATIONS \& FUTURE WORK}
Our work operates under the scientific constraint of data scarcity, using physical priors to enable zero-shot generalization from synthetic data.
This framing naturally informs the primary avenues for future research.

Future work could explore adapting RamPINN for cases where it could learn from a large corpus of unpaired real-world CARS spectra while being fine-tuned on only partially available Raman spectra.

A natural extension of this work would be to design flexible background models, for example, using a learned function before capturing more complex NRB structures.
Similarly, the current architecture assumes a fixed spectral resolution.
Future work could address this by incorporating principles from Neural Operators~\parencite{kovachki2021neural}, yielding a resolution-agnostic model that can adapt to data from different experimental setups without retraining.

\section{CONCLUSION}

In this work, we introduce RamPINN, a physics-informed neural network for reconstructing Raman spectra from CARS signals.
By incorporating physical principles, specifically the Kramers-Kronig relations and NRB smoothness, as physics-based loss terms into the learning process, we demonstrate that RamPINN significantly enhances performance over purely data-driven baselines.
While supervised training naturally yields the best results, the strong inductive bias induced by the physics terms provides competitive recovery even in the fully self-supervised setting.

Beyond spectroscopy, this work highlights a broader opportunity: many scientific domains possess deep, well-established knowledge that can be formally integrated into machine learning.
Rather than replacing domain understanding, combining physical principles with modern learning methods offers a practical and interpretable path forward, especially in scientific settings where data is scarce or synthetic.

Sutton's "bitter lesson" suggests that, given enough data and computing power, learning outperforms inductive hand-crafted biases~\parencite{sutton2019bitter}.
However, in science-driven fields, data is often limited, and the inductive biases we embed are not heuristic approximations, but grounded in physical law.
In these cases, physics-informed learning is not a compromise, but a necessary and principled foundation for generalization.

\acknowledgments{
This study was partially supported by the European Union’s Horizon Europe research and innovation program for the uCAIR project under Grant Agreement No. 101135175.
}

\printbibliography
\clearpage
\appendix
\thispagestyle{empty}

\onecolumn
\aistatstitle{Supplementary Material}

\section{Raman Spectroscopy and CARS}
\label{ap:raman}
Raman spectroscopy is a non-destructive chemical analysis technique widely used for molecular characterization, including in medical and biological applications. 
When light interacts with matter, most photons are elastically scattered (Rayleigh scattering), but a small fraction undergoes inelastic scattering, exchanging energy with molecular vibrational modes. 
This inelastic process produces a shift in photon energy that is characteristic of the molecule’s internal structure~\parencite{raman1928New}.

The resulting Raman spectrum can be interpreted as a high-dimensional feature vector (intensity vs. frequency shift), which is fingerprint to a molecule.
Each peak corresponds to a specific vibrational or rotational transition, enabling identification of chemical composition, molecular structure, orientation, and concentration.
From a data perspective, Raman measurements are typically represented as one-dimensional spectra, where each sample consists of intensity values sampled over a discrete set of wavenumbers.
The main drawback is that the acquisition time in Spontaneous Raman (SR) measurement is very high.
This occurs because, naturally, these interactions are very low, and photon accumulation significant enough for quantification takes time.

In spontaneous Raman (SR) spectroscopy, a monochromatic laser illuminates the sample, and the scattered light is collected and spectrally resolved using a dispersive element (e.g., a grating spectrometer) and a detector such as a CCD. 
Because the Raman scattering cross-section is very small, only a tiny fraction of incident photons contribute to the signal. 
As a result, long acquisition times or signal averaging are required to achieve an acceptable signal-to-noise ratio (SNR). 
This leads to practical limitations such as low throughput, motion sensitivity, and challenges in real-time or in vivo measurements~\parencite{raman_slow2023}.

To overcome this, a faster approach called coherent anti-Stokes Raman scattering (CARS) was introduced~\parencite{bcars9, cars_speed2004}. 
In CARS, at least two synchronized laser beams are used: a pump beam at frequency $\omega_p$ and a Stokes beam at frequency $\omega_s$. 
When the frequency difference $\omega_p - \omega_s$ matches a molecular vibrational resonance, a coherent vibrational excitation is generated in the sample. 
A third interaction with the pump field then produces an anti-Stokes signal at frequency $2\omega_p - \omega_s$.
This uses a coherent process to stimulate the Raman process at a higher rate, such that more photons are released.

Unlike SR, where scattering events are independent and incoherent, CARS is a coherent process: the emitted photons are phase-matched and constructively interfere, leading to a signal that scales quadratically with the number of oscillators. 
This results in significantly higher signal intensity and enables much faster acquisition, often allowing imaging at video rates. 
From a measurement standpoint, CARS systems typically use scanning microscopy setups, where spectra are acquired either point-by-point (spectral scanning) or via multiplex detection schemes.

For machine learning applications, this distinction has important implications. 
SR data is generally linear in the molecular concentration and directly interpretable, whereas CARS signals are proportional to the square modulus of the third-order nonlinear susceptibility, $|\chi^{(3)}|^2$ as seen in \autoref{eq:start} of \autoref{ap:kk}. 
This introduces nonlinear mixing of resonant and non-resonant contributions, meaning that the measured spectrum is not a simple linear superposition of individual molecular signatures. 
Consequently, the inverse problem recovering chemically meaningful spectra from CARS measurements is more complex and often requires preprocessing, modeling, or learning-based approaches.

\newpage

\section{Kramers-Kroning Principle} \label{ap:kk}
The Kramers-Kronig relations, or KK relations, are mathematical relations that connect real and imaginary parts of any complex function that is analytic in the upper half-plane.

In CARS, the NRB originates from electronic contributions to the third-order nonlinear susceptibility ($\chi^{(3)}$)~\parencite{kk_book,kramers1928Diffusion,kronig1926dispersion,guenther2004Encyclopedia}:
\begin{equation}\label{eq:start}
  I_\text{CARS}(\omega ) \propto \left|\chi^{(3)}(\omega )\right|^{2} I_{pu}^{2} I_{s}\,,
\end{equation}

where $I_{pu}$ and $I_{s}$ are the pump and stokes laser intensity, respectively.
The nonlinear susceptibility includes the Raman resonant part $\chi_{r}$  and the non-resonant part $\chi_{nr}$, corresponding to electronic contributions and NRB, respectively.
The non-resonant part is frequency independent.
Excluding the third order in susceptibility notation, this is expressed as:
\begin{equation}\label{eq:split}
  \left|\chi(\omega )\right|^{2} = \left|\chi_{r}(\omega ) + \chi_{nr}\right|^{2} = \left|\chi_{r}(\omega )\right|^{2} + 2\chi_{nr}\Re[\chi_{r}(\omega )] + \left|\chi_{nr}\right|^{2}\,.
\end{equation}

The non-resonant part arises due to the measurement setup, and the resonant part can be mathematically expressed as:
\begin{equation}\label{eq:chi}
  \chi_{r}(\omega) = \sum_{r} \frac{A_{r}}{\Omega_{r}-(\omega_{pu} - \omega_{s}) - i\gamma_{r}}
\end{equation}

where $(\omega_{pu} - \omega_{s})$ is the difference in pump and stokes frequency and $A_{r}$, $\Omega_{r}$ and $\gamma_{r}$ are the amplitude, vibrational frequency and bandwidth of $r^\text{th}$ Raman mode respectively.
We can relate the real and imaginary parts in \autoref{eq:chi} as:
\begin{equation}\label{eq:ph}
  \chi_{r}(\omega)= \left| \chi_{r}(\omega) \right| e^{i \phi(\omega)}\,.
\end{equation}

As $I_\text{Raman}\approx\Im [\chi_{r}(\omega)]$, we need to estimate the phase term~\parencite{kk1}.
From \autoref{eq:start} and \autoref{eq:split}, the CARS measurement only has information on the squared modulus of susceptibility $I_\text{CARS} \approx \left|\chi(\omega)\right|^{2}$.
To approximate NRB, a reference spectrum with Raman inactive substance with susceptibility $\chi_{nr}(\omega)$ is measured and applied such that $\chi_{r}(\omega)= {\chi}/{\chi_{nr}}$.
Now let's represent the normalized intensity as $S(\omega)$. The phase information can be obtained from this using the Kramers-Kronig relations.
It connects the real and imaginary parts as:
\begin{equation}
  \label{eq:kk}
  \begin{aligned}
    \Re [\chi_{r}(\omega)] &= \frac{1}{\pi} \mathcal{P} \int_{-\infty}^{\infty}\frac{\Im [\chi_{r}(\omega')]}{\omega' - \omega}d\omega' \,\text{and}\,\\
    \Im [\chi_{r}(\omega)] &= \frac{1}{\pi} \mathcal{P} \int_{-\infty}^{\infty}\frac{\Re [\chi_{r}(\omega')]}{\omega' - \omega}d\omega' \,.
  \end{aligned}
\end{equation}

Here, the $\mathcal{P}$ represents the Cauchy principal value used to handle improper integrals with singularities by avoiding the singularity through a limiting process.
Using the KK relation\parencite{ph_kk}, we can deduce the phase as:
\begin{equation} \label{kk}
  \varphi (\omega) =
  \frac{1}{\pi} \mathcal{P} \int_{-\infty}^{\infty}\frac{ln(\sqrt{S(\omega')})}{\omega' - \omega}d\omega'\,.
\end{equation}

The Hilbert transform is a specific type of KK relation that applies to real-valued signals.
For a real-valued function $x(t)$:
\begin{equation}
  \label{eq:kk-real}
  \mathcal{H}\{x(t)\} = \frac{1}{\pi} \mathcal{P} \int_{-\infty}^{\infty}\frac{x(t')}{t' - t}dt'\,.
\end{equation}

Therefore, the discrete Hilbert transform on $ln(\sqrt{S(\omega)})$ will give the phase.

\newpage

\section{Synthetic Data Generation}
\label{sec:dg}
To train RamPINN, we generate synthetic CARS spectra by simulating both the resonant Raman signal and the non-resonant background (NRB).
This synthetic dataset is constructed to closely mimic realistic spectral profiles observed in practical CARS experiments. %
Below, we describe the components and procedure used.

\paragraph{Resonant Raman Signal.}
Raman-active modes are modeled as Lorentzian-shaped peaks, following the standard formulation for the third-order nonlinear susceptibility as in~\autoref{eq:chi}. In implementation, the Raman susceptibility $ \chi^{(3)} $ is constructed as:
\begin{equation}
  \chi_r^{(3)}(\omega) = \sum_{r=1}^{N} \frac{A_r}{\Omega_r - \omega - i\gamma_r},
\end{equation}

where:
\begin{itemize}
  \item $ N \sim \mathcal{U}\{1, 25\} $ is number of peaks varied per sample,
  \item $ A_r \in \mathcal{U}(0.01, 1.0) $ is the amplitude (randomized),
  \item $ \Omega_r \in \mathcal{U}(0, 1) $ is the normalized resonance frequency,
  \item $ \gamma_r \in \mathcal{U}(0.001, 0.02) $ is the linewidth,
  \item $ \omega $ is the normalized Raman shift ranging 1000 points over [0, 1].
\end{itemize}

The real and imaginary parts of $\chi^{(3)}$ are retained to compute the CARS intensity and use in the Kramers-Kronig loss. The signal is normalized such that the maximum absolute value of $ \chi^{(3)} $ is 1.

Multiple such Lorentzian peaks are summed to construct a complete Raman spectrum.
The real and imaginary parts of $\chi^{(3)}$ are retained to compute the CARS intensity and use in the Kramers-Kronig loss.

\paragraph{Non-Resonant Background (NRB).}
To simulate the NRB, we use two classes of functions commonly observed in experimental setups:
Sigmoid-based and polynomial-shaped backgrounds.
Each spectrum is randomly assigned one of these forms.

\paragraph{Sigmoid-Based NRB.}
The sigmoid NRB is generated using the following function, where we use a product of two sigmoids to better capture smooth peak-shaped NRB profiles:
\begin{equation}
  \label{eq:double_sigmoid_nrb}
  \chi_{nrb}^{(3)}(\omega) = \frac{1}{1 + e^{-b_1(\omega - c_1)}} \cdot \frac{1}{1 + e^{b_2(\omega - c_2)}}\,,
\end{equation}
where:
\begin{itemize}
  \item $b_1, b_2 \sim \mathcal{N}(10, 5)$ (steepness),
  \item $c_1, c_2 \sim \mathcal{N}(0.2, 0.3)$ and $\mathcal{N}(0.7, 0.3)$ respectively (centers).
\end{itemize}

This form captures the smooth, sigmoidal nature of NRB often observed in real data, as suggested in prior studies~\parencite{polnrb}.

\paragraph{Polynomial-Based NRB.}
Alternatively, NRB is modeled as a smooth polynomial function:
\begin{equation}
  \label{eq:poly_nrb}
  \chi_{nrb}^{(3)}(\omega) = a\omega^4 + b\omega^3 + c\omega^2 + d\omega + e\,,
\end{equation}
with randomly sampled coefficients:
\begin{equation}
  a, c \sim \mathcal{U}(-1, 1), \quad b, d, e \sim \mathcal{U}(-10, 10)\,.
\end{equation}
These coefficients produce a range of smooth background shapes, consistent with observations from experimental NRB profiles.

\paragraph{CARS Spectrum Generation.}
The final CARS spectrum is computed as:
\begin{equation}
  I_{\text{CARS}}(\omega) \propto \left| \chi_{r}^{(3)}(\omega) + \chi_{\text{nrb}} \right|^2\,,
\end{equation}
where $ \chi_{\text{nrb}} $ is treated as real-valued, either polynomial or sigmoid in form, depending on the random draw.
The total spectrum is normalized to ensure that learning focuses on the spectral shape rather than absolute intensity. A random noise $\sim \mathcal{U}(0.0005, 0.003)$ is also added to simulate high frequency disturbances.

\section{RamPINN Model Architecture}
\label{sec:rampinn-model-architecture}

The RamPINN model utilizes a 1D U-Net-like~\parencite{ronneberger2015unet} architecture adapted for spectral data processing.
It takes a single-channel 1D CARS spectrum as input and outputs two separate 1D spectra corresponding to the estimated resonant Raman component and the Non-Resonant Background (NRB) component.
The architecture consists of an encoder path, a bottleneck, and two parallel decoder paths.
We implement the model in PyTorch~\parencite{paszke2019pytorch}.

\paragraph{Encoder.}
The encoder comprises four stages.
Each stage consists of a \texttt{ConvBlock1D} module followed by an \texttt{AvgPool1d} operation with a kernel size of 2 (effectively downsampling by a factor of 2).
The \texttt{ConvBlock1D} module contains a 1D Convolution (\texttt{nn.Conv1d}) with a kernel size of $5$ and padding of $2$ (maintaining sequence length within the block), followed by \texttt{BatchNorm1d} and a \texttt{ReLU} activation.
The number of channels increases through the encoder: $1 (\text{input}) \to 64 \to 128 \to 256 \to 512$.

\paragraph{Bottleneck.}
The bottleneck connects the encoder and decoder paths at the lowest spatial resolution (highest feature abstraction).
It consists of a \texttt{ConvBlock1D}, followed by a \texttt{SelfAttention1D} module, and another \texttt{ConvBlock1D}.
The self-attention mechanism (SelfAttention1D) employs standard scaled dot-product attention, allowing the model to capture long-range dependencies within the compressed spectral representation ($512$ channels).

\paragraph{Decoders.}
Two identical, parallel decoder branches are used to reconstruct the resonant and NRB components separately.
Each decoder consists of four stages, mirroring the encoder structure.
Each stage uses an \texttt{UpBlock1D} module, which first performs linear upsampling (\texttt{nn.Upsample} with scale factor=2), followed by a 1D Convolution (kernel size $5$, padding $2$), \texttt{BatchNorm1d}, and \texttt{ReLU}.
Skip connections are implemented by concatenating (\texttt{torch.cat}) the output of each decoder stage with the feature map from the corresponding encoder stage (after interpolating the decoder feature map to match the encoder feature map's spatial dimension using \texttt{F.interpolate}).
The channel dimensions decrease through the decoder: $512 \to 256 \to 128 \to 64 \to 32$.

\paragraph{Output Layers.}
Each decoder branch terminates with a final 1D Convolution (\texttt{nn.Conv1d} with kernel size $1$) projecting the 32-channel feature map to a single channel.
This output is passed through a \texttt{Sigmoid} activation function, scaling the predicted values between 0 and 1.
Finally, both the resonant and NRB outputs are interpolated (\texttt{F.interpolate} with mode='linear') back to the original input sequence length to ensure compatibility with the ground truth spectra during training and evaluation.

This dual-decoder architecture with shared encoder and bottleneck allows the model to learn common features relevant to both components while dedicating specific pathways for reconstructing the distinct resonant and NRB signals, potentially leveraging shared information from the CARS input effectively.

\subsection{Architecture Selection}
RamPINN model mostly focus on physics loss but an optimal architecture is also trivial for performance.
From \Cref{tab:architecture} we see that the U-Net architecture outperforms all other forms of backends, and that the attention block positively impacts the fidelity of the reconstructed signal (PSNR).
\begin{table}[ht!]
\centering
\begin{adjustbox}{max width=\linewidth}
\begin{tabular}{lcc}
  \toprule
  \textbf{RamPINN Architecture} & \textbf{Supervised} & \textbf{Self-Supervised} \\
  \midrule
  U-Net w/ Attention (Ours)          &\textBF{0.0006 (33.83)}  &\textBF{0.0053 (22.79)} \\
  U-Net w/o Attention                & 0.0007 (31.56)          & 0.0058 (22.59) \\
  ResNet                             & 0.0016 (27.90)          & 0.0078 (21.06) \\
  AutoEncoder                        & 0.0025 (25.96)          & 0.0071 (21.47) \\
  Fourier Neural Operator            & 0.0040 (23.97)          & 0.0135 (18.86) \\
  \bottomrule
\end{tabular}
\end{adjustbox}
\caption{Comparison of different RamPINN backend architectures under supervised and self-supervised training.}
\label{tab:architecture}
\end{table}

\section{Baseline Implementations}
\label{sec:baselines}

To benchmark the performance of RamPINN, we implemented several state-of-the-art deep learning models from literature for CARS-to-Raman signal reconstruction.

\subsection{SpecNet Model}
\label{sec:specnet_baseline}
SpecNet ~\parencite{valensise2020specnet} employs a CNN trained on synthetic CARS data to extract the imaginary part of the third-order susceptibility, corresponding to the Raman signal.
While SpecNet performs well on synthetic datasets, its sensitivity to noise limits its applicability to real-world data, especially in scenarios with weak Raman peaks.
Nonetheless, it provides a valuable benchmark for assessing the impact of noise on model performance. We trained the provided SpecNet till 10 epochs with a learning rate of 0.001, batch size 256 and loss metric as mean square error.

\subsection{LSTM Model}
\label{sec:cnn_lstm_baseline}
\parencite{lstm} introduced LSTM networks to process CARS spectra.
Being a type of recurrent neural network, the LSTM layers model sequential dependencies across the spectrum.
We trained the provided LSTM network till 30 epochs with a learning rate of 0.005, batch size 10 and loss metric as mean absolute error.

\subsection{VECTOR: Very Deep Convolutional Autoencoder Model}
\label{sec:vector_baseline}
The VECTOR model~\parencite{wang2021vector}, is a deep convolutional autoencoder designed to retrieve Raman-like spectra from CARS measurements without requiring NRB reference data.
The architecture comprises an encoder-decoder structure with skip connections to preserve essential spectral features.
Trained on simulated noisy CARS spectra, VECTOR effectively learns to denoise and reconstruct the underlying Raman signal using a customized loss function consisting $L1$ penalty terms.
Its performance surpasses earlier models like SpecNet and LSTM.
We trained the provided VECTOR network till 40 epochs with a learning rate of 0.0001 and batch size 10.

\subsection{BiLSTM Model}
\label{sec:bi_lstm_baseline}
Building upon the LSTM approach, a Bi-LSTM architecture~\parencite{bilstm} for NRB removal in CARS spectra is explored.
The BiLSTM processes the spectral data in both forward and backward directions, capturing comprehensive contextual information.
This model exhibited superior performance in reconstructing Raman signals, particularly in regions with overlapping peaks or complex NRB structures.
We trained the provided Bi-LSTM network till 10 epochs with a learning rate of 0.005, batch size 10 and loss metric as mean squared error.

\subsection{GAN Model}
A GAN model~\parencite{gan} to extract Raman signals from CARS spectra in an adversarial fashion.
The generator in the network has an encoder-decoder architecture with skip connections, and the discriminator is a CNN. We trained the provided GAN model for 1000 epochs with a learning rate of 0.0001 and batch size 8. A customized loss function was employed, utilizing a weighted Huber loss function based on whether peaks are classified as true-positive, false-positive, true-negative, or false-negative, as specified by the authors, for a chosen threshold.

\subsection{CNN+GRU Hybrid Model}
In the same paper~\parencite{gan} proposed a CNN+GRU model that combines a set of convolutional layers and bidirectional gated recurrent units (GRU).
This combination of these models helps to capture long-term dependencies in data.
We trained the provided CNN+GRU network till 50 epochs with a learning rate of 0.0005, batch size 256, and loss metric as Huber loss.

\subsection{TDKK Method}
Time-Domain Kramers-Kronig (TDKK)~\parencite{kk_nrb1} is a traditional algorithm for NRB removal in broadband CARS microscopy, utilizing phase retrieval techniques. While effective, it requires additional NRB measurement with a buffer sample, and estimations of hyperparameters are necessary.

\subsection{LeDHT Method}
A learned-matrix representation of the discrete Hilbert transform (LeDHT) method~\parencite{ledht} introduces a learned matrix approach to the discrete Hilbert transform, enhancing the accuracy of Raman signal extraction by addressing distortions caused by NRB interference. It is computationally efficient and significantly enhances the performance of traditional phase retrieval techniques, such as the Kramers–Kronig relation. The approach involves training a matrix that approximates the Hilbert transform, which is then applied to new spectra via matrix multiplication.

\subsection{IWT Method}
IWP \parencite{interp_wp} introduces an approach that utilizes wavelet-based decomposition to model and correct additive or multiplicative spectral background errors. This method is implemented in MATLAB and can also be applied in an unsupervised setting, with the drawback of incomplete phase removal.

\newpage
\subsection{Model Card Overview}

We provide a short overview of model parameters and differences for all baseline models and our proposed RamPINN architecture.
The values are provided in \autoref{tab:model_comparison}. We estimate model size using each framework’s built-in parameter-counting tools.

\begin{table}[h]
  \centering
  \resizebox{\textwidth}{!}{%
    \begin{tabular}{l c c c c c c c}
      \toprule
      & SpecNet & VECTOR & BiLSTM & LSTM & CNN+GRU & GAN & RamPINN (Ours) \\
      \midrule
      \# Parameters  & 6.0M & 111.8M & 51.4k & 3.9k & 84.0k & 6.2M & 6.8M \\
      \midrule
      Architecture & CNN & AE & BiLSTM & LSTM & \makecell{CNN +\\BiGRU} & \makecell{Gen (AE) \\ Disc (CNN)} & \makecell{AE + \\ Attention} \\
      \midrule
      Loss Function & L2 & L1 & L2 & L1 & Huber & Adversarial &
      \makecell{Physics-\\Informed} \\
      \midrule
      Framework & TensorFlow & PyTorch & TensorFlow & TensorFlow & TensorFlow & PyTorch & PyTorch \\
      \bottomrule
    \end{tabular}%
  }
  \caption{Comparison of model parameters, architectures, and loss functions. "AE" denotes Autoencoder; "Gen" Generator; "Disc" Discriminator.} %
  \label{tab:model_comparison}
\end{table}

\newpage
\FloatBarrier
\section{Kramers--Kronig Implementation}
\label{sec:kramers--kronig-implementation}

\paragraph{Physical Intuition.}
The KK relations originate from causality in linear response theory:
for any physical system, the imaginary (absorptive) part of the susceptibility is determined by the real (dispersive) part via the Hilbert transform.
In the context of CARS spectroscopy, the measured signal combines a broadband non-resonant background (NRB) and a sharp resonant term.
The NRB contributes primarily to the real part of the susceptibility.
By estimating and removing the NRB, we isolate a residual that, under ideal conditions, should behave as the real part of a causal analytic signal.
The imaginary part of this analytic continuation, obtained via the Hilbert transform, corresponds to the true Raman signature.
The KK loss thus enforces this structure by minimizing the discrepancy between the predicted Raman component and the imaginary part of the analytic residual.

\FloatBarrier
\subsection{RamPINN Optimization Algorithm}
\label{ap:algorithm-training}
\begin{algorithm}
  \centering
  \begin{algorithmic}[1]
\Require CARS spectra $x$, optional Raman labels $y_{\text{true}}$, learning rate $\eta$, number of iterations $T$
\Ensure Trained network $y_\theta(x) = (y_{\text{res}}, y_{\text{NRB}})$

\For{$t = 1$ \textbf{to} $T$}
    \State Sample mini-batch $x$
    \State $(y_{\text{res}}, y_{\text{NRB}}) \gets y_\theta(x)$
    \State $r \gets x - y_{\text{NRB}}$ %
    
    \State \textbf{Hilbert Transform:}
    \State $R_f \gets \text{FFT}(r)$
    \State Construct $H_f$ %
    \State $\tilde{R}_f \gets R_f \cdot H_f$
    \State $z \gets \text{IFFT}(\tilde{R}_f)$
    \State $y_{\text{res}}^{(\text{KK})} \gets \Im(z)$

    \State $\loss_{\text{KK}} \gets \| y_{\text{res}} - y_{\text{res}}^{(\text{KK})} \|^2$
    \State $\loss_{\text{smooth}} \gets \| \nabla y_{\text{NRB}} \|^2$

    \If{$y_{\text{true}}$ is available}
        \State $\loss_{\text{MSE}} \gets \| y_{\text{res}} - y_{\text{true}} \|^2$
    \Else
        \State $\loss_{\text{MSE}} \gets 0$
    \EndIf

    \State $\loss_{\text{total}} \gets \lambda_{\text{KK}} \cdot \loss_{\text{KK}} + \lambda_{\text{smooth}} \cdot \loss_{\text{smooth}} + \lambda_{\text{data}} \cdot \loss_{\text{MSE}}$
    \State $\theta \gets \theta - \eta \cdot \nabla_\theta \loss_{\text{total}}$
\EndFor
\end{algorithmic}

  \label{al:}
\end{algorithm}

\FloatBarrier
\newpage

\section{Additional Synthetic Samples}
\label{ap:reconstruction-plots}

\begin{figure}[ht]
  \centering
  \includegraphics[width=0.98\linewidth]{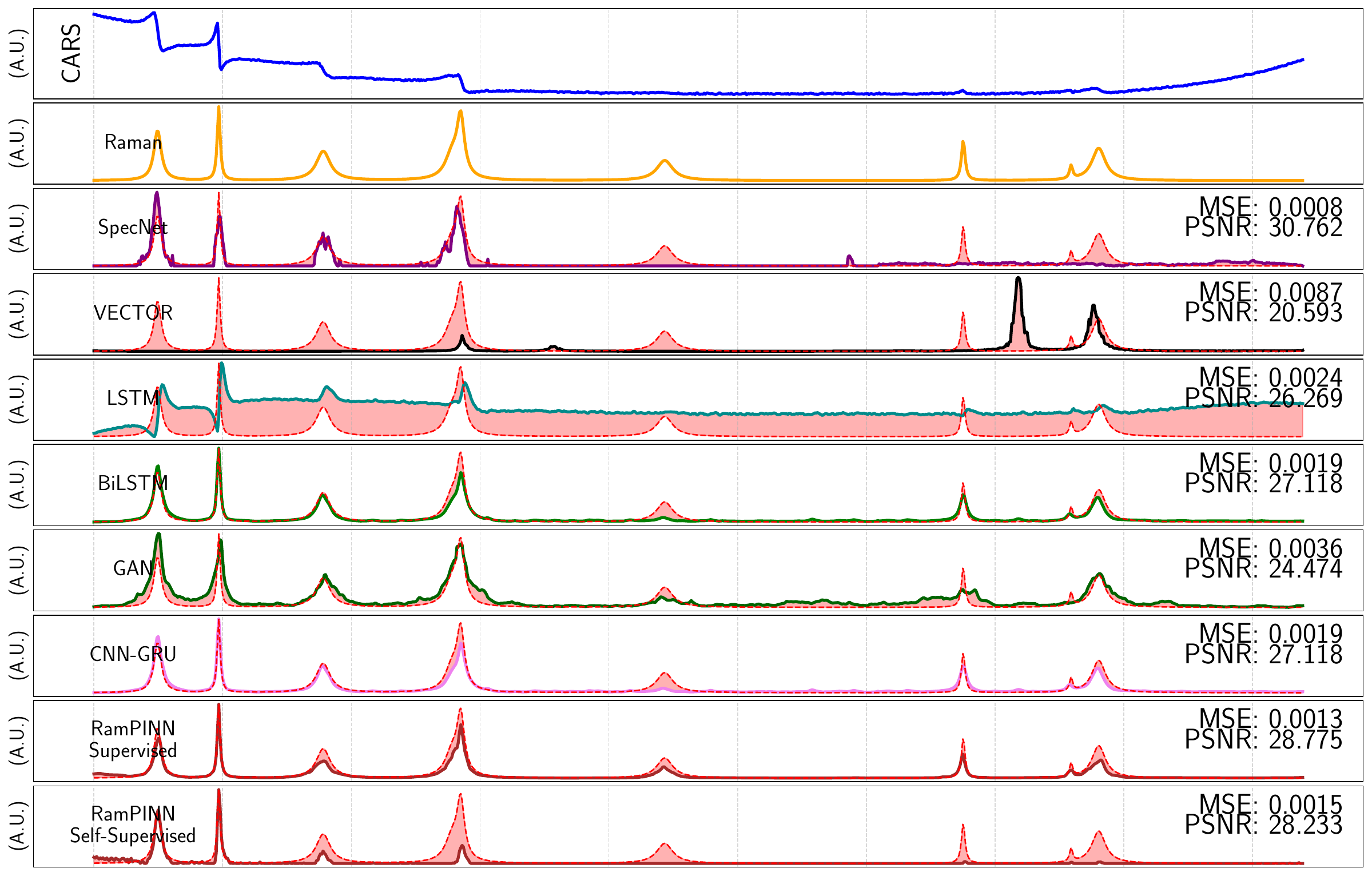}
  \caption{
    \textbf{Additional qualitative comparison of Raman reconstruction.}
  }
  \label{fig:app:vis-001}
\end{figure}

\begin{figure}[ht]
  \centering
  \includegraphics[width=0.98\linewidth]{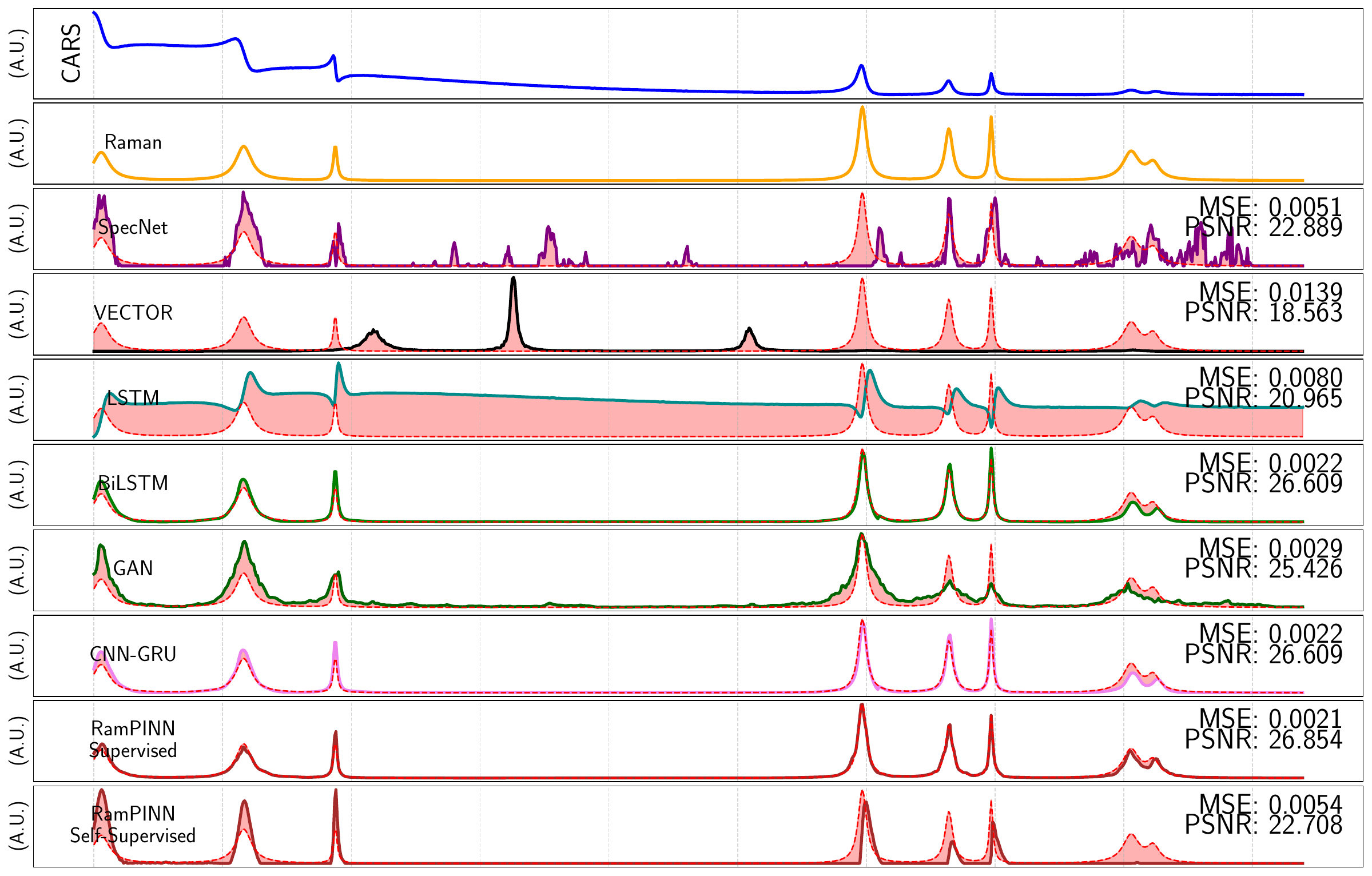}
  \caption{
    \textbf{Additional qualitative comparison of Raman reconstruction.}
  }
  \label{fig:app:vis-002}
\end{figure}

\begin{figure}[ht]
  \centering
  \includegraphics[width=0.98\linewidth]{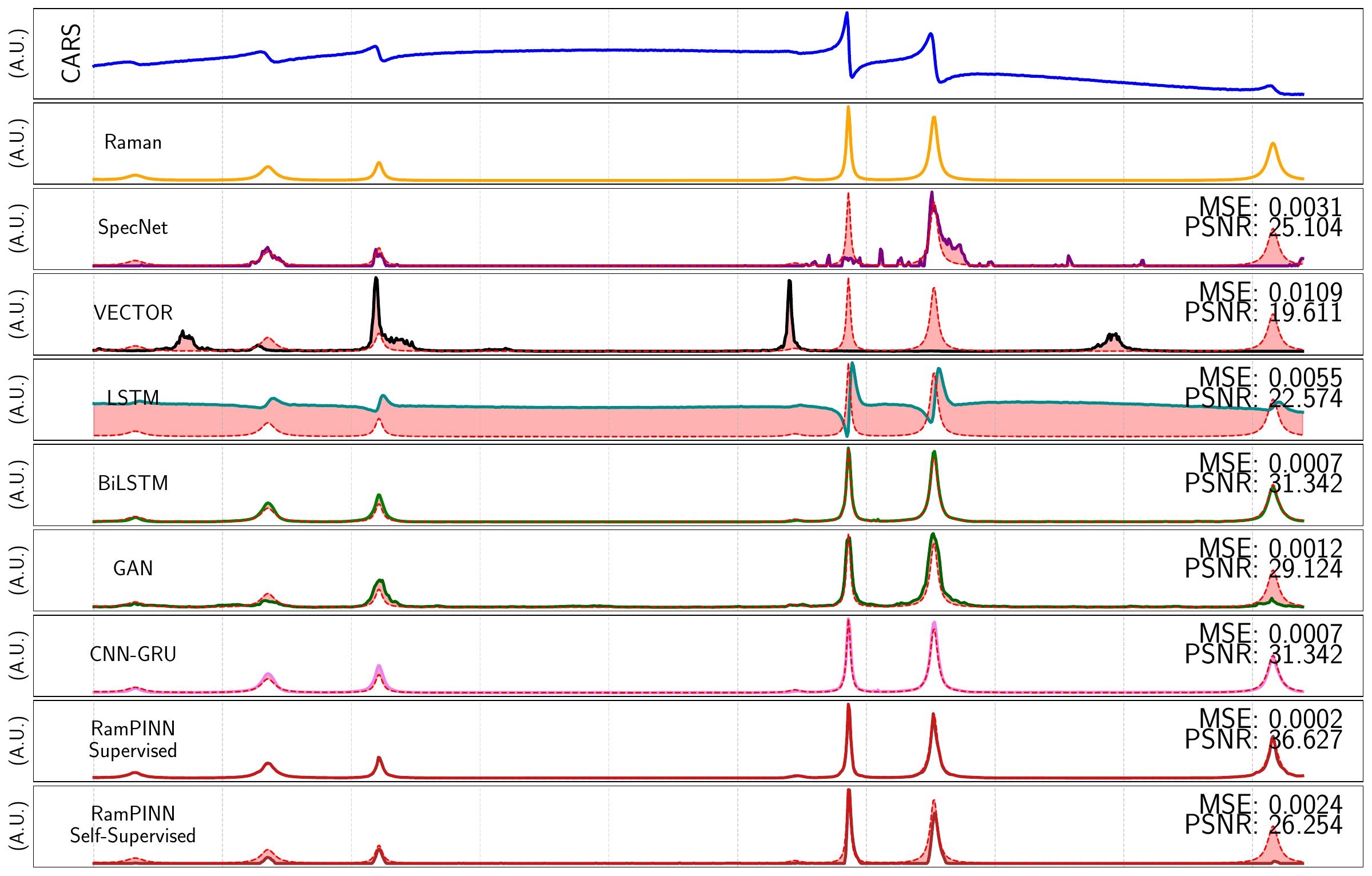}
  \caption{
    \textbf{Additional qualitative comparison of Raman reconstruction.}
  }
  \label{fig:app:vis-012}
\end{figure}

\begin{figure}[ht]
  \centering
  \includegraphics[width=0.98\linewidth]{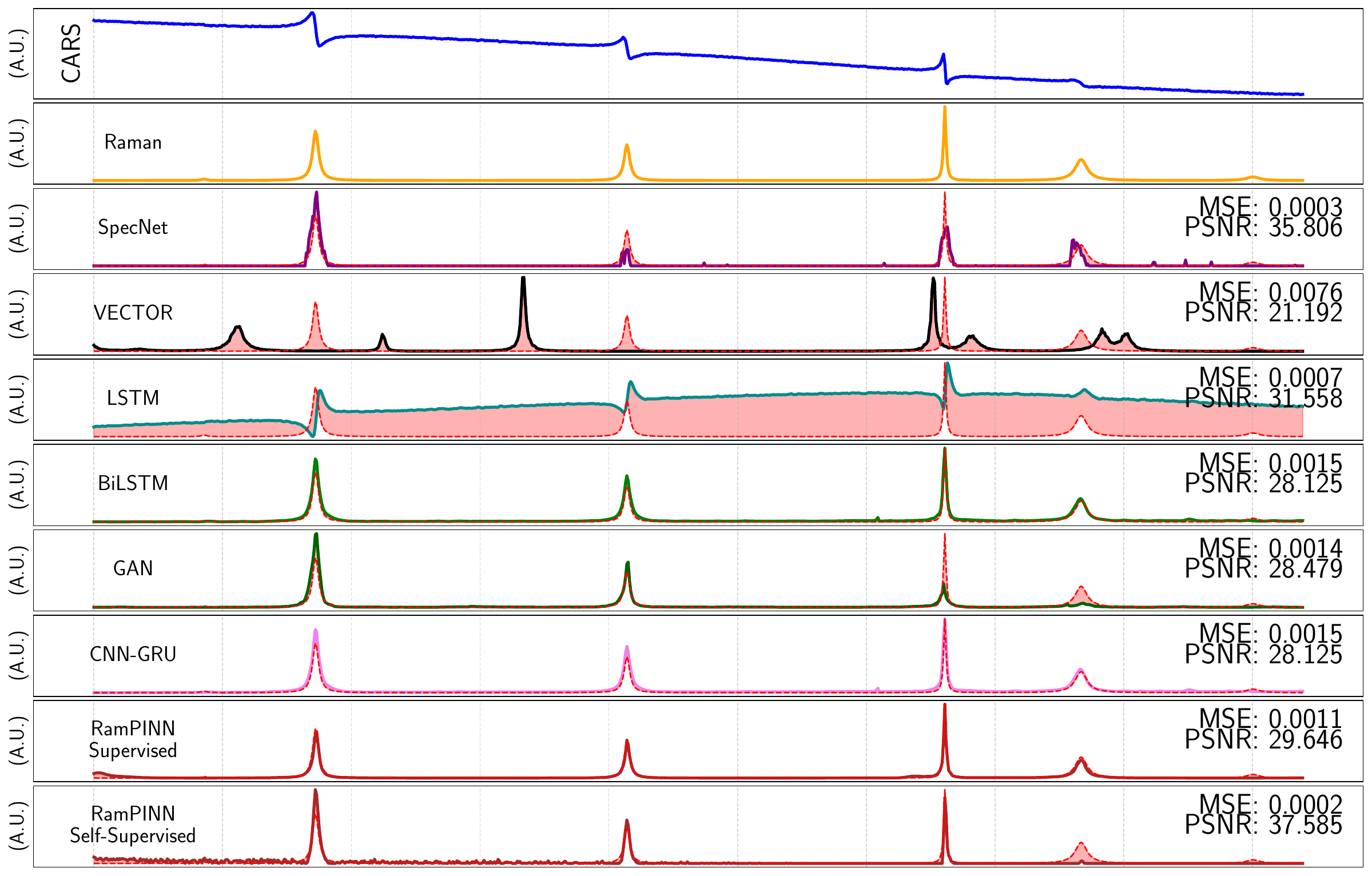}
  \caption{
    \textbf{Additional qualitative comparison of Raman reconstruction.}
  }
  \label{fig:app:vis-356}
\end{figure}

\FloatBarrier
\newpage
\section{Additional Real-World Samples}
\label{sec:additional-real-world-samples}

\begin{figure}[ht]
  \centering
  \includegraphics[width=\textwidth]{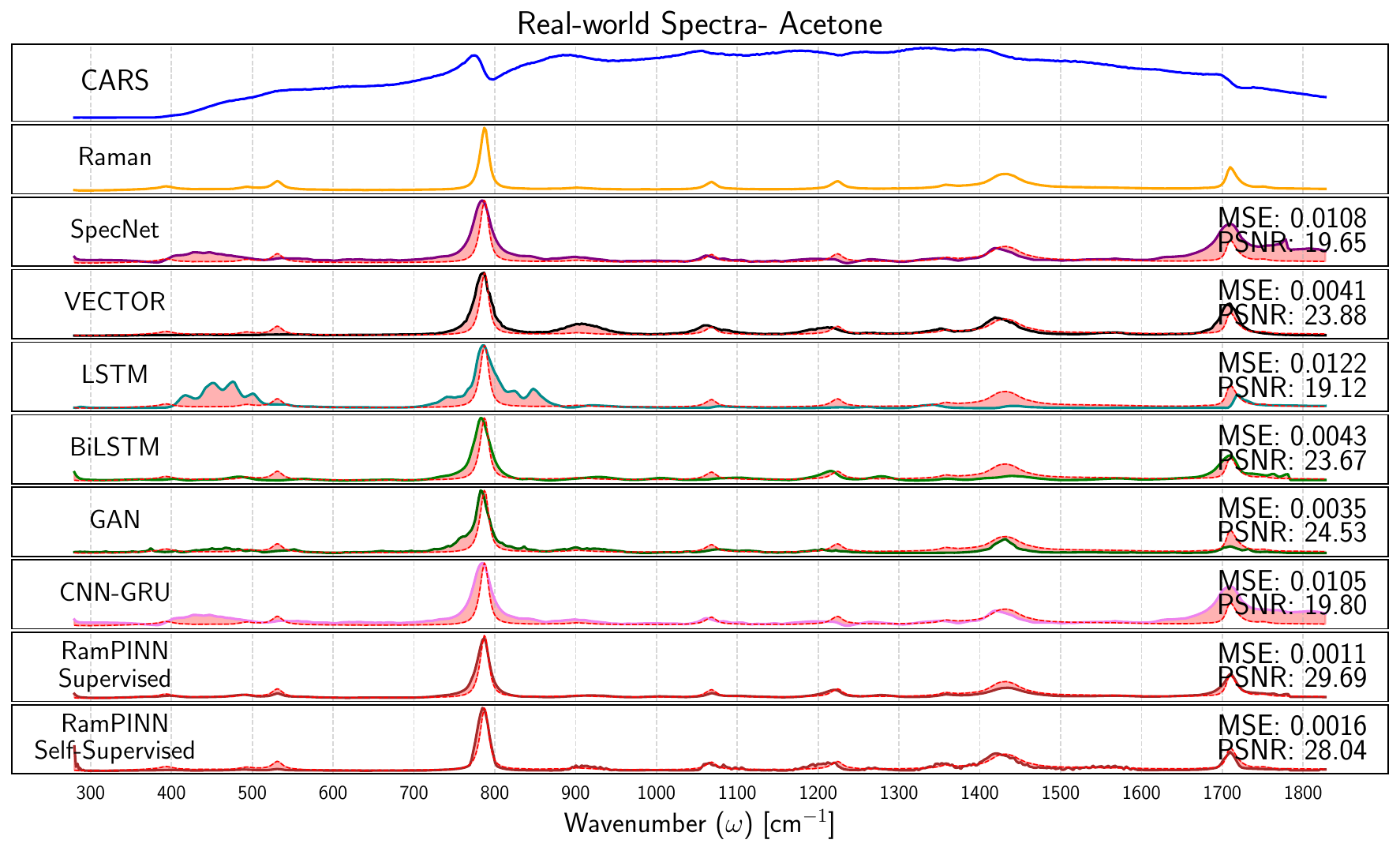}
  \caption{\textbf{Real-world sample -- Acetone.}}
  \label{fig:supp-realworld-acetone}
\end{figure}

\begin{figure}[ht]
  \centering
  \includegraphics[width=\textwidth]{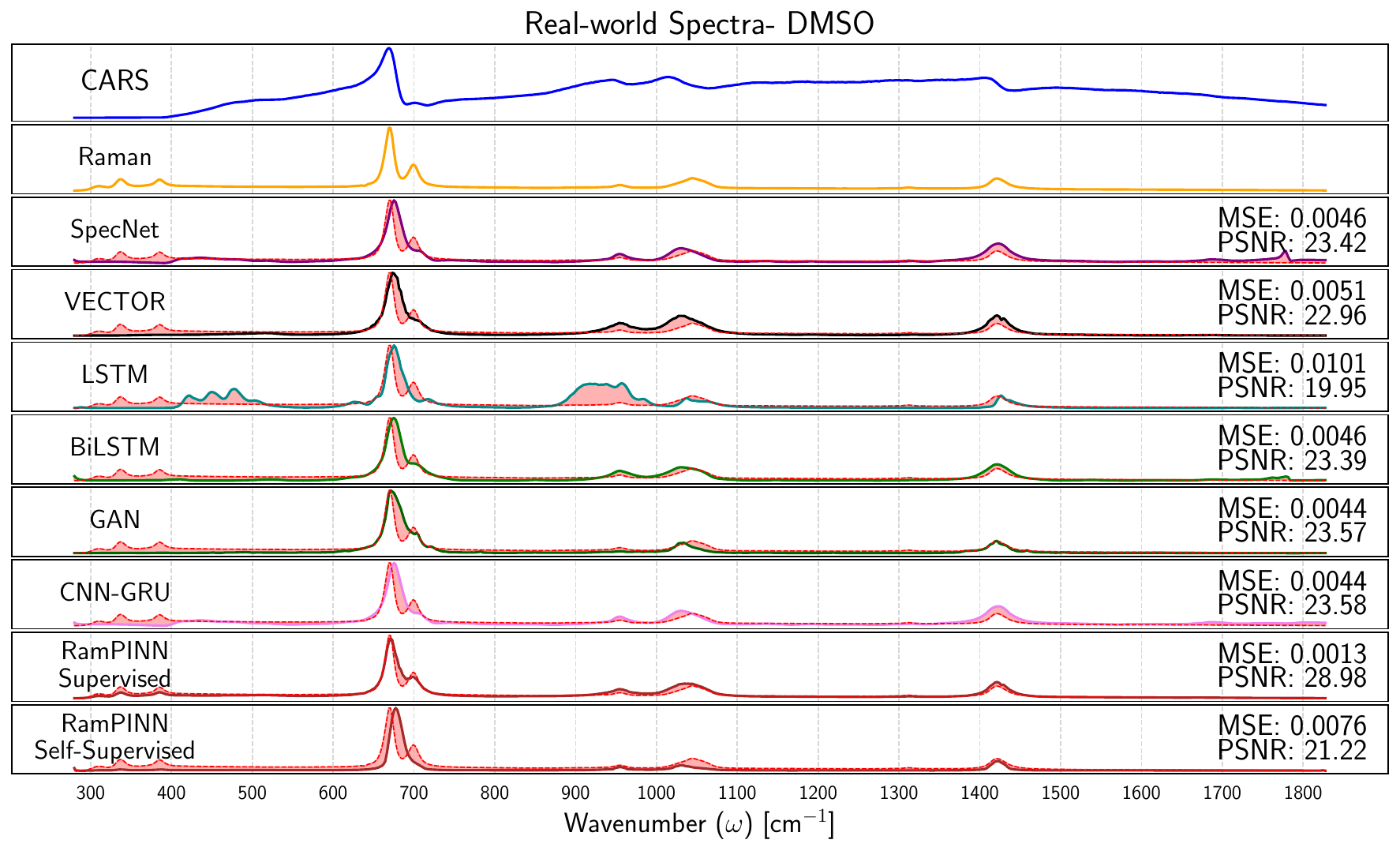}
  \caption{\textbf{Real-world sample -- DMSO.}}
  \label{fig:supp-realworld-dmso}
\end{figure}

\begin{figure}[ht]
  \centering
  \includegraphics[width=\textwidth]{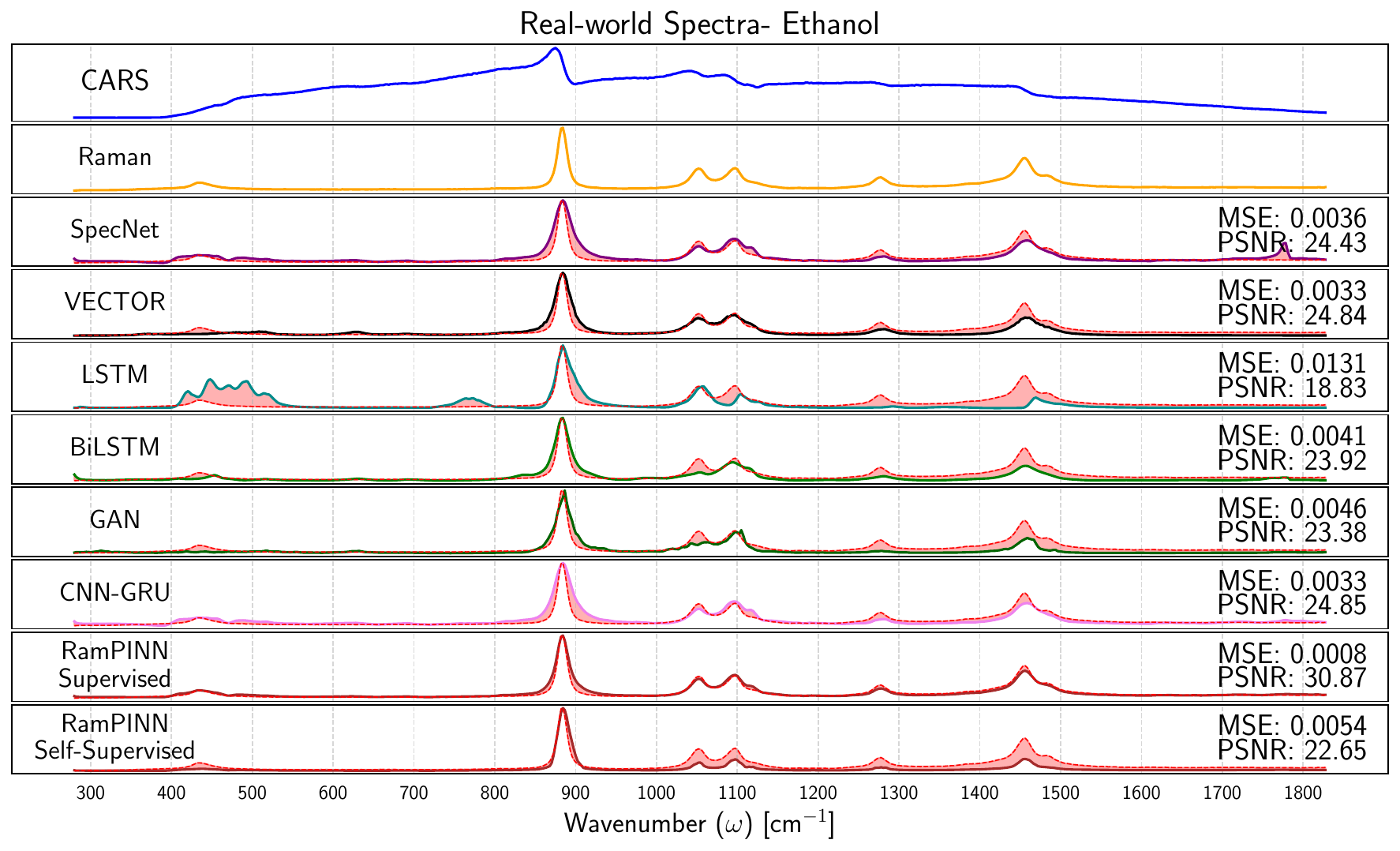}
  \caption{\textbf{Real-world sample -- Ethanol.}}
  \label{fig:supp-realworld-ethanol}
\end{figure}

\begin{figure}[ht]
  \centering
  \includegraphics[width=\textwidth]{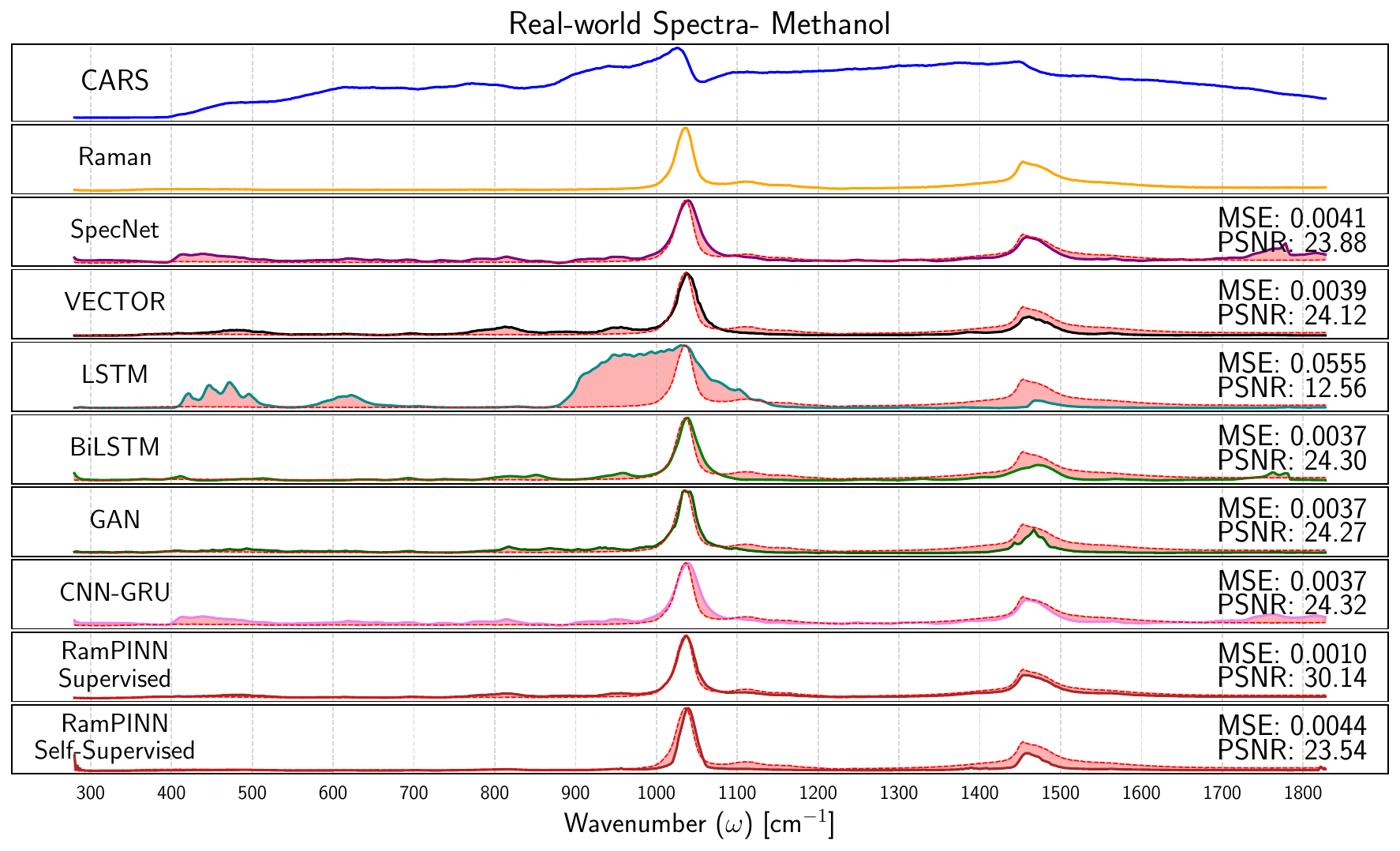}
  \caption{\textbf{Real-world sample -- Methanol.}}
  \label{fig:supp-realworld-methanol}
\end{figure}

\begin{figure}[ht]
  \centering
  \includegraphics[width=\textwidth]{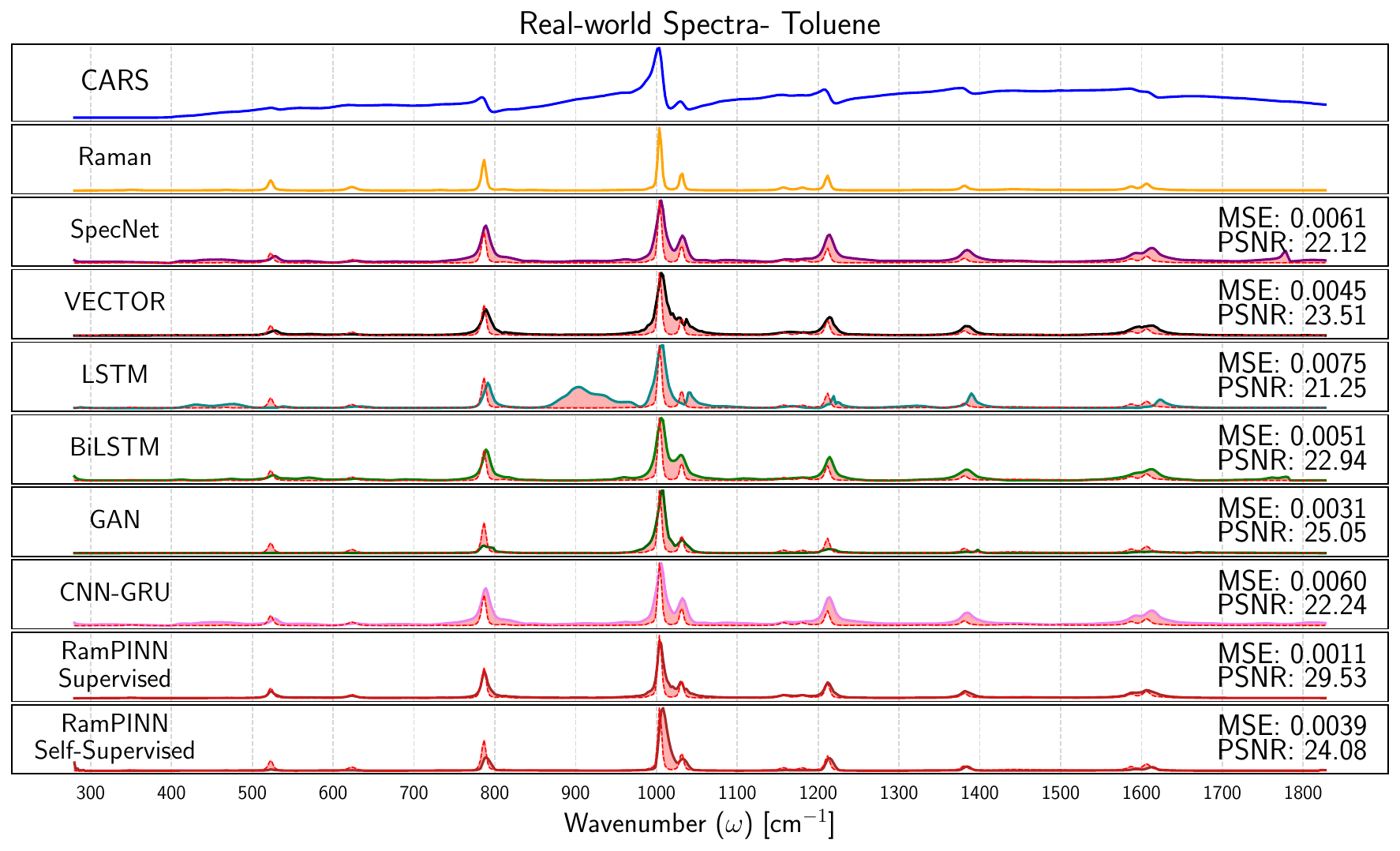}
  \caption{\textbf{Real-world sample -- Toluene.}}
  \label{fig:supp-realworld-toluene}
\end{figure}

\begin{figure}[ht]
  \centering
  \includegraphics[width=\textwidth]{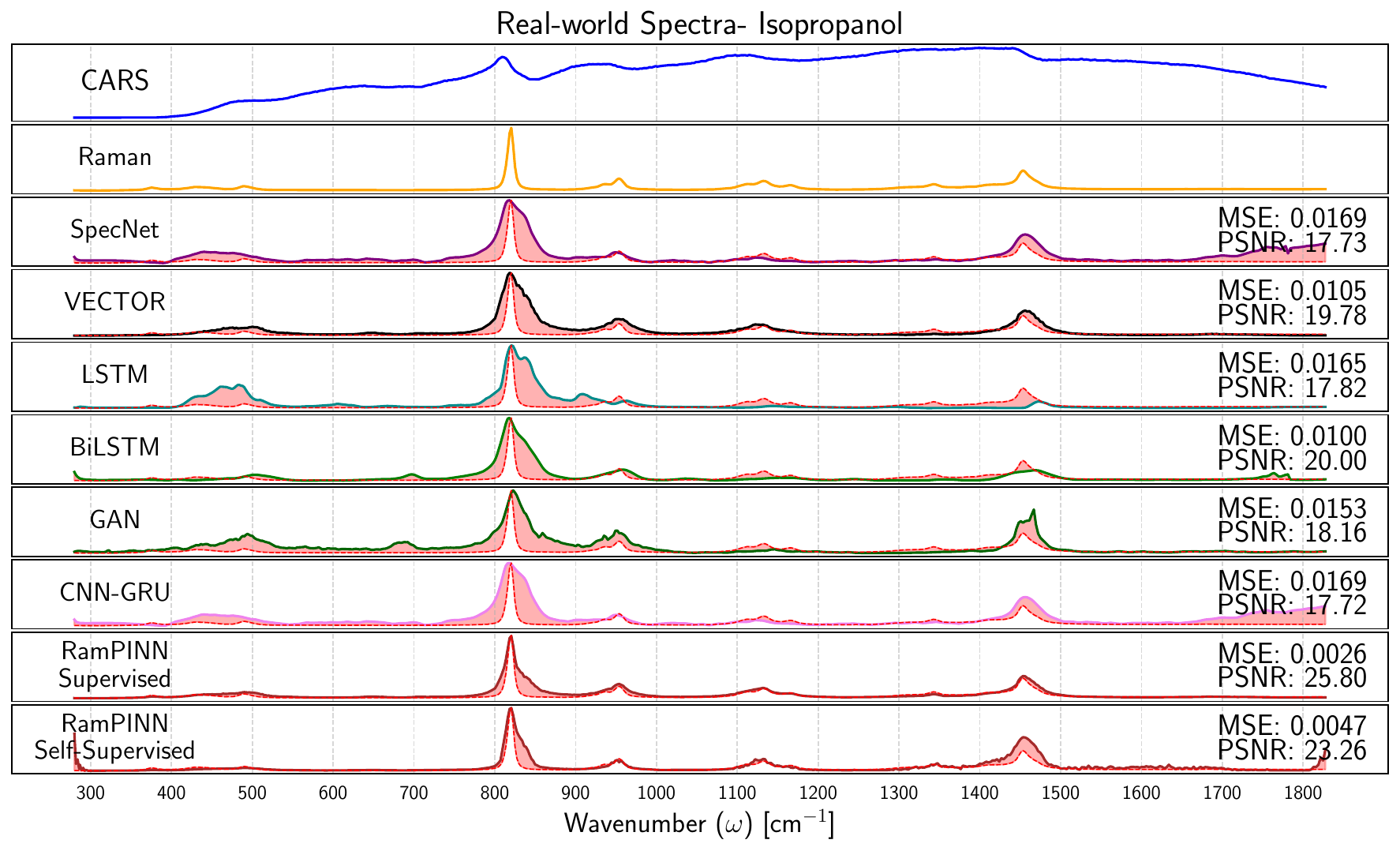}
  \caption{\textbf{Real-world sample -- Isopropanol.}}
  \label{fig:supp-realworld-isopropanol}
\end{figure}
\FloatBarrier

\newpage

\section{Robustness Studies}
\label{ap:rob}
Our physics-informed objective is guided by the prior that the non-resonant background (NRB) is a smooth, peak-free component.
While this is a common and valid approximation, real-world experimental conditions can introduce complexities.
Therefore, we conducted two targeted experiments to assess the robustness of RamPINN when these core assumptions are violated.

\paragraph{Robustness to Increased Nonlinearity.}
To test the model's ability to handle more complex background shapes, we evaluated its performance on synthetic data with increasingly nonlinear NRB profiles.
We generated these by increasing the maximum polynomial degree used in the simulation (see Appendix C for details).
\Cref{fig:rob_nl} shows that the model's performance exhibits graceful degradation.
Even with highly complex, nonlinear backgrounds, the reconstruction error remains low, demonstrating that the model is not brittle and can robustly handle a wide range of smooth background shapes.
This indicates that the smoothness prior acts as a robust inductive bias, guiding the model to prefer smoother solutions rather than forcing it to fit only simple ones.

\paragraph{Robustness to Peak-like Artifacts.}
To simulate sharp instrumental artifacts or other features that violate the peak-free assumption, we injected a varying number of random Gaussian peaks into the input CARS spectra.
These peaks, intentionally distinct from the Lorentzian shapes of true Raman signals, had randomly sampled heights.
The results, presented in \Cref{fig:rob_peak}, show that RamPINN is remarkably resilient to such structured artifacts.
The model effectively learns to attribute these sharp, non-physical features to the NRB component, thereby preserving the fidelity of the reconstructed Raman signal.
This demonstrates the efficacy of our physics-architecture co-design: the network learns that physically-inconsistent features are best explained by the NRB decoder, allowing it to isolate the true Raman signal even in the presence of strong, peak-like interference.

\begin{figure*}[ht!]
  \centering
  \begin{subfigure}[b]{0.48\textwidth}
    \centering        \includegraphics[width=\linewidth]{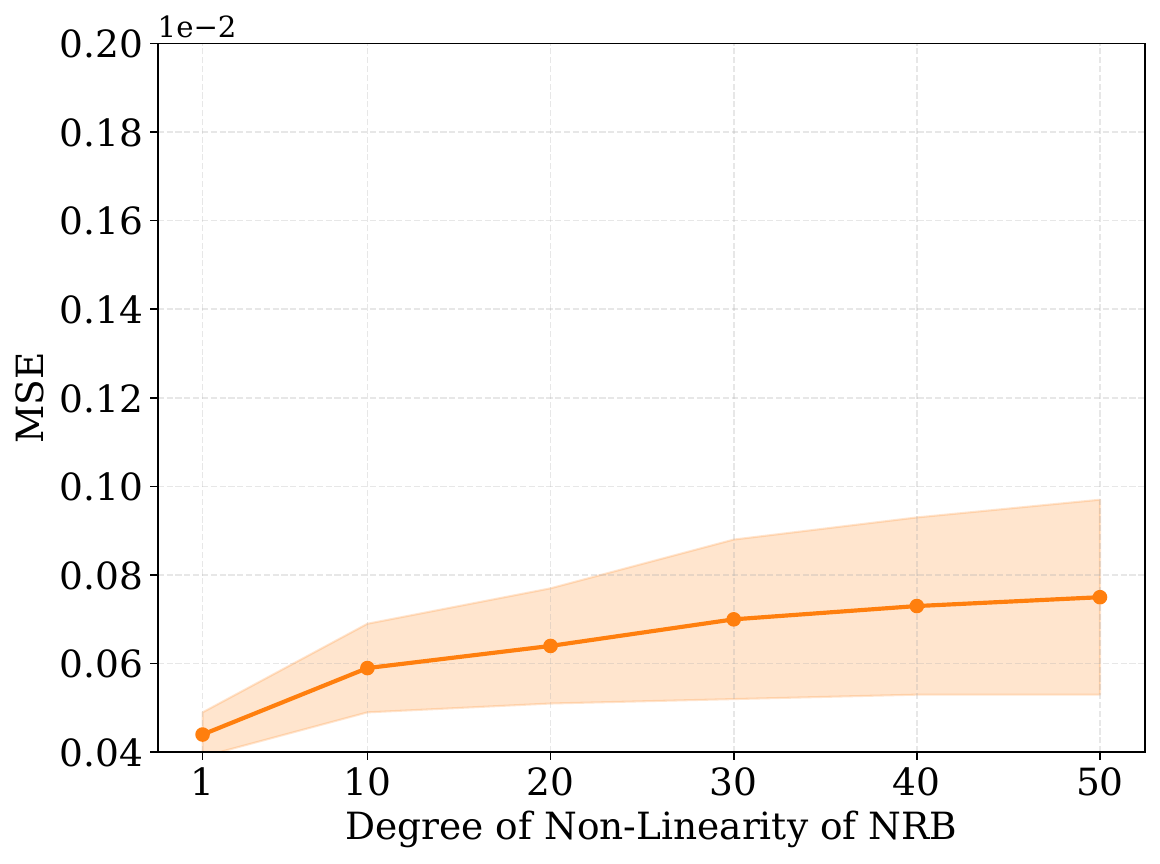}
    \caption{
      Robustness to increasing nonlinearity in NRB.
    }
    \label{fig:rob_nl}
  \end{subfigure}
  \hfill
  \begin{subfigure}[b]{0.48\textwidth}
    \centering
    \includegraphics[width=\linewidth]{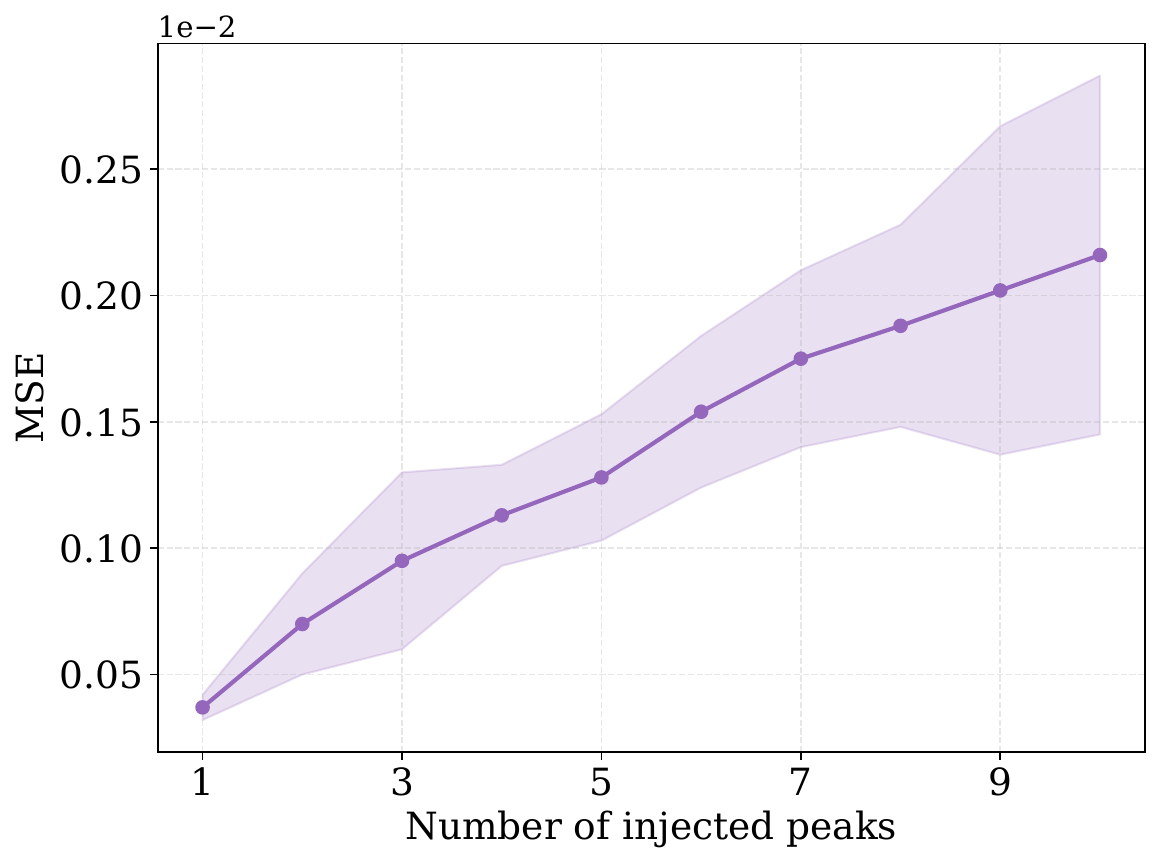}
    \caption{
      Robustness to random peaks injected into CARS.
    }
    \label{fig:rob_peak}
  \end{subfigure}
  \hfill
  \caption{
    \textbf{Robustness of RamPINN}.
  }
  \label{fig:ablations_appendix}
\end{figure*}

\FloatBarrier
\newpage

\section{Additional Metrics}
\label{app:metrics}

MSE and PSNR summarize pointwise differences and are useful for broad comparisons, but they do not reflect spectroscopic goals.
In Raman spectra tasks, what matters is whether the peaks are present, where they are, and how strong they are.
Two typical failure modes illustrate the gap:
\begin{enumerate}

  \item A  small wavenumber shift of a narrow peak can be chemically significant yet only slightly affect MSE;
  \item Smooth background deviations can inflate MSE while the peak set (locations and intensities) is essentially correct.
\end{enumerate}

\paragraph{Setting and detection.}
All spectra are normalized in intensity. Positions are evaluated on a normalized axis
$x = i/(L-1) \in [0,1]$.
We detect peaks with \texttt{scipy.signal.find\_peaks} on an optionally smoothed signal, using absolute (scale-free) thresholds under normalization:
minimum height $h_{\min} \in [0,1]$, minimum prominence $p_{\min} \in [0,1]$, and a minimum separation
$\delta \in (0,1]$ expressed as a fraction of the signal length.
Predicted and true peaks are matched one-to-one by nearest neighbor within tolerance $\tau$ in normalized units: a match if $|\hat{x} - x| \le \tau$.
Therefore, alongside MSE/PSNR (main paper), we report peak-aware metrics that directly target detection, localization, and intensity fidelity.

\paragraph{Detection Metrics}
We count $\mathrm{TP}$ (predicted peaks matched within $\tau$), $\mathrm{FP}$ (unmatched predicted peaks), and $\mathrm{FN}$ (unmatched true peaks), and compute
\begin{align}
  \mathrm{Precision} &= \frac{\mathrm{TP}}{\mathrm{TP}+\mathrm{FP}}\,,\qquad
  \mathrm{Recall} = \frac{\mathrm{TP}}{\mathrm{TP}+\mathrm{FN}}\,,\\
  \mathrm{F1} &= \frac{2\,\mathrm{Precision}\,\mathrm{Recall}}{\mathrm{Precision}+\mathrm{Recall}}\,.
\end{align}
This measures whether we found the right peaks.

Given $N$ spectra, we provide (a) \emph{macro} averages: mean~$\pm$~std across spectra for each metric (ignoring NaNs for spectra with no matches in location/intensity metrics); (b) \emph{micro} F1: sum $\mathrm{TP},\mathrm{FP},\mathrm{FN}$ over all spectra first, then compute $\mathrm{Precision}$, $\mathrm{Recall}$, and $\mathrm{F1}$ from these totals.

\paragraph{Defaults Parameters.}
Unless stated otherwise, we fix
$\tau=0.01$, $p_{\min}=0.02$, $h_{\min}=0$, $\delta=0.01$, and a Savitzky--Golay window of $11$.
We keep $p_{\min}$, $h_{\min}$, $\delta$ fixed across methods for fairness, and vary only $\tau$ if needed to account for sampling density.

\begin{table}[htbp!]
  \centering
  \caption{
    \textbf{Detection Metrics (Macro and Micro) for the Test Synthetic Dataset.}
    Higher is better ($\uparrow$). Values are mean $\pm$ std over spectra; best per column in \textbf{bold}.
    We observe that RamPINN outperforms in these metrics, demonstrating that it not only retrieves the Raman spectrum effectively but also accurately identifies the peaks.
  }
  \label{tab:detection}
  \resizebox{\textwidth}{!}{
    \begin{tabular}{lcccccc}
      \toprule
      Method & Macro Precision $\uparrow$ & Macro Recall $\uparrow$ & Macro F1 $\uparrow$ & Micro Precision $\uparrow$ & Micro Recall $\uparrow$ & Micro F1 $\uparrow$ \\
      \midrule
      RamPINN   & \textBF{0.983 $\pm$ 0.082} & \textBF{0.972 $\pm$ 0.092} & \textBF{0.960 $\pm$ 0.091} & \textBF{0.986} &\textBF{0.972}  & \textBF{0.962} \\
      RamPINN(self-sup) & 0.925 $\pm$ 0.189 & 0.743 $\pm$ 0.225 & 0.797 $\pm$ 0.192 & 0.917 & 0.710 & 0.800 \\
      BiLSTM    & 0.852 $\pm$ 0.174 & \textBF{0.972} $\pm$\textBF{0.092} & 0.896 $\pm$ 0.129 & 0.878 & 0.962 & 0.923 \\
      CNN-GRU   & 0.852 $\pm$ 0.174 & 0.945 $\pm$ 0.116 & 0.896 $\pm$ 0.129 & 0.878 & 0.940 & 0.923 \\
      SpecNet   & 0.260 $\pm$ 0.142 & 0.813 $\pm$ 0.196 & 0.370 $\pm$ 0.158 & 0.241 & 0.789 & 0.369 \\
      VECTOR    & 0.114 $\pm$ 0.150 & 0.131 $\pm$ 0.179 & 0.106 $\pm$ 0.120 & 0.116 & 0.127 & 0.121 \\
      GAN       & 0.029 $\pm$ 0.115 & 0.020 $\pm$ 0.077 & 0.020 $\pm$ 0.068 & 0.092 & 0.020 & 0.032 \\
      \bottomrule
    \end{tabular}
  }
\end{table}

\paragraph{Mean Location Error (Normalized).}
Over matched pairs $(\hat{x}_j, x_j)$,
\begin{equation}
  \mathrm{MLE}_{\mathrm{norm}} = \frac{1}{\mathrm{TP}}\sum_{j=1}^{\mathrm{TP}}\lvert \hat{x}_j - x_j \rvert \in [0,1]\,,
\end{equation}
reported as NaN if $\mathrm{TP}=0$.
This measures how well peak positions align.

\paragraph{Relative Intensity Error.}
Let $A$ denote the per-peak amplitude. For each match,
\begin{equation}
  r_j = \frac{\lvert \hat{A}_j - A_j \rvert}{\max\!\bigl(\lvert A_j \rvert,\varepsilon\bigr)}\,.
\end{equation}
We report the mean and median of $\{r_j\}$; NaN if $\mathrm{TP}=0$. This measures how well peak magnitudes are recovered.

\begin{table}[hbtp!]
  \centering
  \caption{
    \textbf{Localization and intensity errors on test synthetic data.}
    Lower is better ($\downarrow$). Values are mean $\pm$ std; best per column in \textbf{bold}. We demonstrate that RamPINN outperforms classical methods in these metrics, indicating that RamPINN effectively aligns peak positions and magnitudes.
  }
  \label{tab:errors}
  \resizebox{\textwidth}{!}{
    \begin{tabular}{lccccc}
      \toprule
      Method & Mean loc. err. (norm) $\downarrow$ & Rel. int. err. (mean) $\downarrow$ & Rel. int. err. (median) $\downarrow$ & Pooled RIE (mean) $\downarrow$ & Pooled RIE (median) $\downarrow$ \\
      \midrule
      RamPINN  & \textbf{0.0004 $\pm$ 0.0004} & \textbf{0.399 $\pm$ 0.254} & \textbf{0.329 $\pm$ 0.181} & \textbf{0.429} & \textbf{0.309} \\
      RamPINN(self-sup) & 0.0009 $\pm$ 0.0008 & 0.449 $\pm$ 0.167 & 0.447 $\pm$ 0.196 & 0.455 & 0.453 \\
      \midrule
      BiLSTM    & \textbf{0.0004 $\pm$ 0.0004} & 0.674 $\pm$ 0.319 & 0.650 $\pm$ 0.319 & 0.705 & 0.641 \\
      CNN-GRU   & 0.0020 $\pm$ 0.0011 & 0.674 $\pm$ 0.319 & 0.650 $\pm$ 0.319 & 0.705 & 0.641 \\
      SpecNet   & 0.0006 $\pm$ 0.0004 & 0.599 $\pm$ 0.336 & 0.581 $\pm$ 0.341 & 0.599 & 0.525 \\
      VECTOR    & 0.0049 $\pm$ 0.0026 & 2.238 $\pm$ 3.733 & 2.122 $\pm$ 3.643 & 2.271 & 0.812 \\
      GAN       & 0.0046 $\pm$ 0.0024 & 0.857 $\pm$ 1.075 & 0.859 $\pm$ 1.075 & 0.824 & 0.816 \\
      \bottomrule
    \end{tabular}
  }
\end{table}

\paragraph{Zero-shot Evaluation on Real Molecules.}
We applied our detection-focused metrics—precision, recall, F1, relative intensity error (mean; $\downarrow$), and mean location error (normalized; $\downarrow$)—to six real-world molecules (acetone, DMSO, ethanol, methanol, toluene, isopropanol); see Tables~\Cref{tab:acetone,tab:dmso,tab:ethanol,tab:methanol,tab:toluene,tab:isopropanol}.
These metrics directly assess whether a model finds the \emph{right peaks} and aligns them at the \emph{right locations} with the \emph{right intensities}, complementing global reconstruction scores like MSE/PSNR.

\noindent\textbf{Takeaway.}
Not only does \textbf{RamPINN} reconstruct signals well (MSE/PSNR), it also \emph{identifies} chemically plausible peaks and preserves their \emph{alignment} and \emph{relative intensities} under zero-shot distribution shift.
This consistency across detection and alignment metrics evidences the strength of our physics-integrated approach.

\begin{table}[hbtp!]
  \centering
  \caption{Acetone: per-model metrics. Arrows indicate direction; best per column in \textbf{bold}.}
  \label{tab:acetone}
  \resizebox{\textwidth}{!}{
    \begin{tabular}{lccccc}
      \toprule
      Method & Precision $\uparrow$ & Recall $\uparrow$ & F1 $\uparrow$ & Rel. Int. Err. (mean) $\downarrow$ & Mean Loc. Err. (norm) $\downarrow$ \\
      \midrule
      RamPINN   & 0.748 & \textbf{0.875} & \textbf{0.778} & \textbf{0.251} & \textbf{0.0009} \\
      RamPINN(self-sup) & 0.489 & \textbf{0.875} & 0.538 & 0.600 & 0.0026 \\
      \midrule
      BiLSTM    & 0.417 & 0.625 & 0.500 & 0.332 & 0.0040 \\
      CNN-GRU   & 0.545 & 0.750 & 0.632 & 0.394 & 0.0036 \\
      VECTOR    & \textbf{0.750} & 0.750 & 0.750 & 1.137 & 0.0036 \\
      SpecNet   & 0.500 & 0.750 & 0.600 & 0.393 & 0.0036 \\
      GAN       & 0.133 & 0.500 & 0.211 & 0.631 & 0.0031 \\
      LSTM      & 0.364 & 0.500 & 0.421 & 0.535 & 0.0038 \\
      \bottomrule
    \end{tabular}
  }
\end{table}

\begin{table}[hbtp!]
  \centering
  \caption{DMSO: per-model metrics. Arrows indicate direction; best per column in \textbf{bold}.}
  \label{tab:dmso}
  \resizebox{\textwidth}{!}{
    \begin{tabular}{lccccc}
      \toprule
      Method & Precision $\uparrow$ & Recall $\uparrow$ & F1 $\uparrow$ & Rel. Int. Err. (mean) $\downarrow$ & Mean Loc. Err. (norm) $\downarrow$ \\
      \midrule
      RamPINN   & \textbf{1.000} & \textbf{1.000} & \textbf{1.000} & 0.093 & 0.0009 \\
      RamPINN(self-sup) & \textbf{1.000} & 0.571 & 0.727 & 0.141 & 0.0038 \\
      \midrule
      BiLSTM    & 0.800 & 0.571 & 0.667 & 0.541 & 0.0031 \\
      CNN-GRU   & 0.667 & 0.571 & 0.615 & 0.615 & 0.0034 \\
      VECTOR    & 0.600 & 0.429 & 0.500 & 0.751 & \textbf{0.0005} \\
      SpecNet   & 0.571 & 0.571 & 0.571 & 0.615 & 0.0034 \\
      GAN       & 0.033 & 0.143 & 0.054 & \textbf{0.090} & 0.0031 \\
      LSTM      & 0.400 & 0.571 & 0.471 & 2.065 & 0.0031 \\
      \bottomrule
    \end{tabular}
  }
\end{table}

\begin{table}[t]
  \centering
  \caption{Ethanol: per-model metrics. Arrows indicate direction; best per column in \textbf{bold}.}
  \label{tab:ethanol}
  \resizebox{\textwidth}{!}{
    \begin{tabular}{lccccc}
      \toprule
      Method & Precision $\uparrow$ & Recall $\uparrow$ & F1 $\uparrow$ & Rel. Int. Err. (mean) $\downarrow$ & Mean Loc. Err. (norm) $\downarrow$ \\
      \midrule
      RamPINN   & \textbf{1.000} & \textbf{1.000} & \textbf{1.000} & \textbf{0.120} & \textbf{0.0003} \\
      RamPINN(self-sup) & \textbf{1.000} & \textbf{1.000} & \textbf{1.000} & 0.502 & 0.0005 \\
      \midrule
      BiLSTM    & 0.625 & 0.833 & 0.714 & 0.407 & 0.0009 \\
      CNN-GRU   & 0.667 & \textbf{1.000} & 0.800 & 0.229 & 0.0010 \\
      VECTOR    & 0.714 & 0.833 & 0.769 & 0.278 & 0.0006 \\
      SpecNet   & 0.667 & \textbf{1.000} & 0.800 & 0.229 & 0.0010 \\
      GAN       & 0.095 & \textbf{1.000} & 0.174 & 0.725 & 0.0036 \\
      LSTM      & 0.455 & 0.833 & 0.588 & 0.563 & 0.0052 \\
      \bottomrule
    \end{tabular}
  }
\end{table}

\begin{table}[t]
  \centering
  \caption{Methanol: per-model metrics. Arrows indicate direction; best per column in \textbf{bold}.}
  \label{tab:methanol}
  \resizebox{\textwidth}{!}{
    \begin{tabular}{lccccc}
      \toprule
      Method & Precision $\uparrow$ & Recall $\uparrow$ & F1 $\uparrow$ & Rel. Int. Err. (mean) $\downarrow$ & Mean Loc. Err. (norm) $\downarrow$ \\
      \midrule
      RamPINN   & \textbf{0.500} & \textbf{0.667} & \textbf{0.571} & 0.032 & \textbf{0.0000} \\
      RamPINN(self-sup) & 0.333 & \textbf{0.667} & 0.444 & 0.185 & 0.0023 \\
      \midrule
      BiLSTM    & 0.125 & 0.333 & 0.182 & \textbf{0.024} & \textbf{0.0000} \\
      CNN-GRU   & 0.200 & \textbf{0.667} & 0.308 & 0.054 & 0.0023 \\
      VECTOR    & 0.400 & \textbf{0.667} & 0.500 & 0.161 & 0.0023 \\
      SpecNet   & 0.200 & \textbf{0.667} & 0.308 & 0.054 & 0.0023 \\
      GAN       & 0.020 & 0.333 & 0.037 & 0.935 & 0.0046 \\
      LSTM      & 0.250 & \textbf{0.667} & 0.364 & 0.369 & 0.0069 \\
      \bottomrule
    \end{tabular}
  }
\end{table}

\begin{table}[t]
  \centering
  \caption{Toluene: per-model metrics. Arrows indicate direction; best per column in \textbf{bold}.}
  \label{tab:toluene}
  \resizebox{\textwidth}{!}{
    \begin{tabular}{lccccc}
      \toprule
      Method & Precision $\uparrow$ & Recall $\uparrow$ & F1 $\uparrow$ & Rel. Int. Err. (mean) $\downarrow$ & Mean Loc. Err. (norm) $\downarrow$ \\
      \midrule
      RamPINN   & \textbf{1.000} & 0.827 & \textbf{0.842} & \textbf{0.283} & \textbf{0.0002} \\
      RamPINN(self-sup) & 0.857 & 0.545 & 0.667 & 0.330 & 0.0020 \\
      \midrule
      BiLSTM    & 0.875 & 0.636 & 0.737 & 0.549 & 0.0020 \\
      CNN-GRU   & 0.889 & 0.727 & 0.800 & 0.568 & 0.0023 \\
      VECTOR    & \textbf{1.000} & 0.636 & 0.778 & 0.470 & 0.0024 \\
      SpecNet   & 0.800 & 0.727 & 0.762 & 0.565 & 0.0023 \\
      GAN       & 0.159 & \textbf{0.909} & 0.270 & 1.008 & 0.0050 \\
      LSTM      & 0.667 & 0.727 & 0.696 & 0.487 & 0.0059 \\
      \bottomrule
    \end{tabular}
  }
\end{table}

\begin{table}[hbtp!]
  \centering
  \caption{Isopropanol: per-model metrics. Arrows indicate direction; best per column in \textbf{bold}.}
  \label{tab:isopropanol}
  \resizebox{\textwidth}{!}{
    \begin{tabular}{lccccc}
      \toprule
      Method & Precision $\uparrow$ & Recall $\uparrow$ & F1 $\uparrow$ & Rel. Int. Err. (mean) $\downarrow$ & Mean Loc. Err. (norm) $\downarrow$ \\
      \midrule
      RamPINN   & \textbf{1.000} & \textbf{0.667} & \textbf{0.800} & \textbf{0.159} & \textbf{0.0000} \\
      RamPINN(self-sup) & 0.375 & \textbf{0.667} & 0.480 & 0.397 & 0.0008 \\
      \midrule
      BiLSTM    & 0.500 & 0.444 & 0.471 & 0.445 & 0.0046 \\
      CNN-GRU   & 0.600 & \textbf{0.667} & 0.632 & 0.956 & 0.0036 \\
      VECTOR    & 0.833 & 0.556 & 0.667 & 0.283 & 0.0034 \\
      SpecNet   & 0.545 & \textbf{0.667} & 0.600 & 0.957 & 0.0036 \\
      GAN       & 0.125 & 0.556 & 0.204 & 0.899 & 0.0061 \\
      LSTM      & 0.444 & 0.444 & 0.444 & 1.524 & 0.0042 \\
      \bottomrule
    \end{tabular}
  }
\end{table}

\end{document}